\documentclass[sigconf,screen]{acmart}
\usepackage{algorithm}

\usepackage{algpseudocode}
\usepackage{makecell}
\usepackage{bm}
\usepackage{multirow}
\usepackage{enumitem}
\usepackage{soul} % 引入soul宏包
\usepackage{xcolor} % 引入xcolor宏包以支持自定义颜色
\usepackage{caption}
\usepackage{subcaption}
\usepackage{hyperref}

\definecolor{lightcyan}{rgb}{0.5, 1, 1}

\AtBeginDocument{%
  }

\copyrightyear{2025}
\acmYear{2025}
\setcopyright{acmlicensed}
\acmDOI{10.1145/3746027.3754502}

\acmConference[MM '25] {Proceedings of the 33rd ACM International Conference on Multimedia}{October 27--31, 2025}{Dublin, Ireland.}
\acmBooktitle{Proceedings of the 33rd ACM International Conference on Multimedia (MM '25), October 27--31, 2025, Dublin, Ireland}
\acmISBN{979-8-4007-2035-2/2025/10}

\acmSubmissionID{60}

\begin{document}

\title{Multi-Object Sketch Animation with Grouping and Motion Trajectory Priors}
%Decoupled Motion Modeling for Multi-Object Sketch Animations via Group-Aware Displacement Networks
%User-Guided Vector Animation Generation: Integrating Motion Trajectory Priors and Interactive Grouping

% \author{Guotao Liang, Juncheng Hu, Ximing Xing, Jing Zhang, Qian Yu}
% \authornote{Corresponding author} 
% \email{{liangguotao, hujuncheng, ximingxing, zhang\_jing, qianyu}@buaa.edu.cn}
% \affiliation{
%   \institution{Beihang University, Beijing, China}
% }
\author{Guotao Liang}
\orcid{0009-0006-0583-7098}
% \affiliation{
%   \department{School of Software \& \\ Qingdao Research Institute}
%   \institution{Beihang University}
%   % \city{Beijing}
%   % \state{Beijing}
%   \country{China}
% }
\affiliation{
  \department{School of Software}
  \institution{Beihang University}
  \city{Beijing}
  \country{China}
}
%\affiliation{
%  \department{Qingdao Research Institute}
%  \institution{Beihang University}
%  \city{Qingdao}
%  \state{Shandong}
%  \country{China}
%}
\email{liangguotao@buaa.edu.cn}

\author{Juncheng Hu}
\orcid{0009-0006-2044-8314}
% \affiliation{
% \department{School of Software}
%   \institution{Beihang University}
%     \city{Beijing}
%   % \state{Beijing}
%   \country{China}
% }
\affiliation{
  \department{School of Software}
  \institution{Beihang University}
  \city{Beijing}
  \country{China}
}
%\affiliation{
%  \department{Qingdao Research Institute}
%  \institution{Beihang University}
%  \city{Qingdao}
%  \state{Shandong}
%  \country{China}
%}
\email{hujuncheng@buaa.edu.cn}

\author{Ximing Xing}
\orcid{0000-0002-7961-4499}
\affiliation{
\department{School of Software}
  \institution{Beihang University}
    \city{Beijing}
  % \state{Beijing}
  \country{China}
}
\email{ximingxing@buaa.edu.cn}

\author{Jing Zhang}
\orcid{0000-0003-3516-0111}
% \affiliation{
% \department{School of Software}
%   \institution{Beihang University}
%     \city{Beijing}
%   % \state{Beijing}
%   \country{China}
% }
\affiliation{
  \department{School of Software}
  \institution{Beihang University}
  \city{Beijing}
  \country{China}
}
\affiliation{
  \department{Qingdao Research Institute}
  \institution{Beihang University}
  \city{Qingdao}
  \state{Shandong}
  \country{China}
}
\email{zhang\_jing@buaa.edu.cn}
\author{Qian Yu}
\orcid{0000-0002-0538-7940}
\authornote{Corresponding author} 
% \affiliation{
%   \department{School of Software \& \\ Qingdao Research Institute}
%   \institution{Beihang University}
%   % \city{Beijing}
%   % \state{Beijing}
%   \country{China}
% }
\affiliation{
  \department{School of Software}
  \institution{Beihang University}
  \city{Beijing}
  \country{China}
}
\affiliation{
  \department{Qingdao Research Institute}
  \institution{Beihang University}
  \city{Qingdao}
  \state{Shandong}
  \country{China}
}
\email{qianyu@buaa.edu.cn}

%\renewcommand{\shortauthors}{Liang et al.}

\begin{abstract}
We introduce GroupSketch, a novel method for vector sketch animation that effectively handles multi-object interactions and complex motions. 
Existing approaches struggle with these scenarios, either being limited to single-object cases or suffering from temporal inconsistency and poor generalization. 
To address these limitations, our method adopts a two-stage pipeline comprising Motion Initialization and Motion Refinement. 
In the first stage, the input sketch is interactively divided into semantic groups and key frames are defined, enabling the generation of a coarse animation via interpolation. 
In the second stage, we propose a Group-based Displacement Network (GDN), which refines the coarse animation by predicting group-specific displacement fields, leveraging priors from a text-to-video model. 
GDN further incorporates specialized modules, such as Context-conditioned Feature Enhancement (CCFE), to improve temporal consistency. 
Extensive experiments demonstrate that our approach significantly outperforms existing methods in generating high-quality, temporally consistent animations for complex, multi-object sketches, thus expanding the practical applications of sketch animation. 
Project page: \href{https://hjc-owo.github.io/GroupSketchProject/}{GroupSketch}.
%Vector animation offers a powerful medium for illustrating the interaction logic of hand-drawn objects within real-world contexts. However, existing techniques typically treat motion within the canvas as a monolithic process, overlooking the physical relationships and inter-dependencies among individual objects. As a result, they struggle to decouple multiple object motions, leading to rigid and uncoordinated interactions. To address these limitations, we propose an interactive framework for generating vector-based multi-object sketch animations. Our approach assigns each object an independent motion trajectory consistent with physical laws, enabling self-consistent and organic interactions among objects. To support flexible animation control, we introduce an intuitive canvas-based interface that allows users to interact, thereby enabling motion trajectory acquisition. Extensive experiments demonstrate that our method significantly outperforms state-of-the-art baselines, especially in scenarios involving multi-object interactions. Our framework achieves superior consistency between sketches and generated videos, as well as higher fidelity in capturing inter-object dynamics, thereby bridging the gap between static illustrations and dynamic, interactive animations.

\end{abstract}

%%
%% Keywords. The author(s) should pick words that accurately describe the work being presented. Separate the keywords with commas.



\maketitle

\section{Introduction}

\begin{figure*}
  \includegraphics[width=\textwidth]{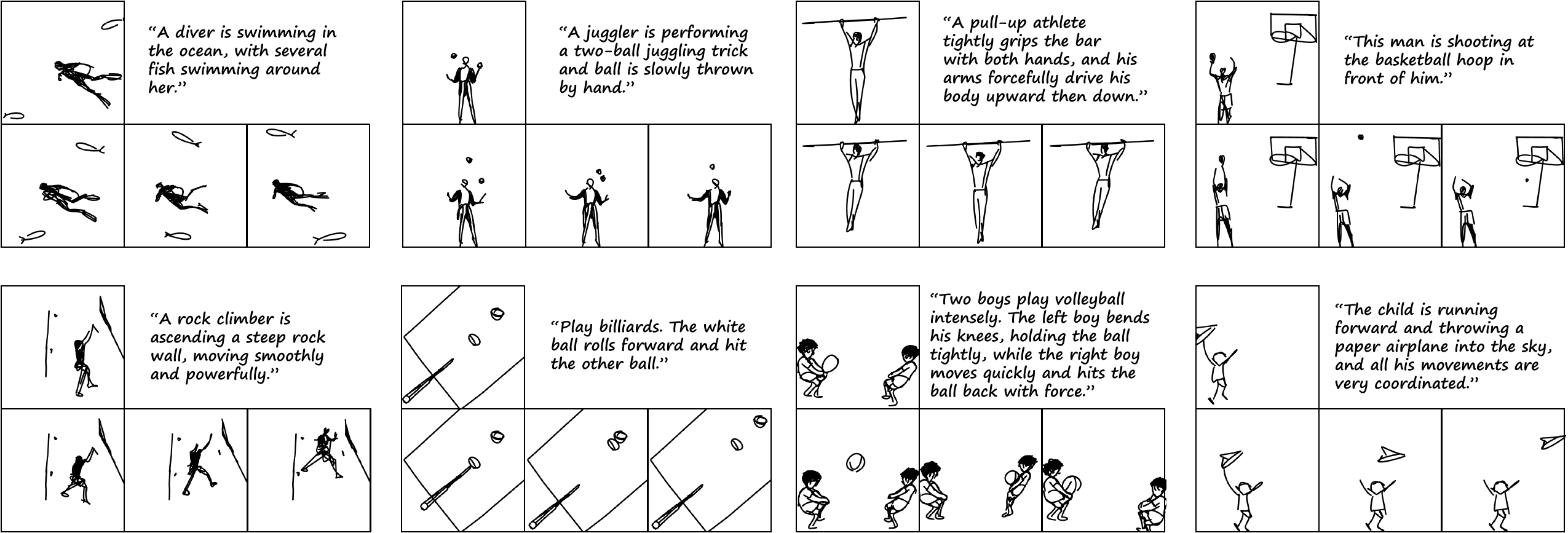}
  % \vspace{-2em}
  \caption{Our model takes a vector sketch and a driving prompt that specifies the desired motion, and generates a short video where the sketch moves in accordance with the given prompt.}
  \label{fig:teaser}
  % \vspace{-1em}
\end{figure*}

%As a basic and instinctive means of visual representation and communication, sketches play a crucial role~\cite{aubert2014pleistocene, fan2023drawing, gombrich1995story}. Sketch animation, which converts static stroke drawings into lively video sequences, is widely employed in storytelling, illustrations, websites, animation production, and entertainment. In this paper, we propose a multi-objective grouping and user-interactive sketch animation generation method.

% why study sketch animation? Logic: what is sketch animation, and its applications --> this task is valuable.
Sketches have been used as important tools for design and expression due to their strong expressivity and intuitive nature. 
With the rapid development of generative models, sketch generation has attracted increasing attention from both academia~\cite{frans2022clipdraw,xing2024diffsketcher,vinker2022clipasso,vinker2023clipascene,vinker2024sketchagent,qu2023sketchdreamer,yu2016sketch,yu2017sketch} and industry~\cite{berger2013style,ha2018a,zheng2024sketch}. Generating dynamic sketches, i.e., sketch animation, is particularly valuable due to its wide range of applications, such as film production, gaming, and education.

% related works: progress and limitations (single object, cannot handle complex motions)--> This task is challenging.
However, animating a sketch remains a significant challenge. 
Two representative approaches have emerged: LiveSketch~\cite{gal2024breathing} and FlipSketch~\cite{bandyopadhyay2024flipsketch}. 
Both methods leverage the motion priors from pre-trained text-to-video (T2V) models~\cite{wang2023modelscope}. 
% TODO：including “Sketch Video Synthesis” ？
LiveSketch animates \textit{vector} sketches by iteratively optimizing the coordinates of the Bezier curves, whereas FlipSketch focuses on animating \textit{raster} sketches by applying DDIM inversion~\cite{song2021ddim} to extract noise patterns, which are then processed by a fine-tuned T2V model. Both of these methods have significantly advanced the field of sketch animation.

Despite this progress, both approaches struggle with complex motions and multi-object scenarios. 
Specifically, LiveSketch is designed for single-object cases and cannot achieve natural motion or interaction in multi-object cases. For example, in animations such as \textit{``a player playing basketball''}, the basketball unnaturally sticks to the player's hand. 
On the other hand, FlipSketch can realize scene-level interactions by directly generating frame sequences. However, it suffers from unexpected changes in appearance and limited temporal consistency due to its raster-based nature.
Moreover, FlipSketch lacks explicit design considerations for multi-object interaction, instead relying on self-constructed training data to learn motion patterns, which potentially limits its generalization ability. These limitations restrict the practical applications of these methods.

% Our method: briefly introduce the pipeline of our method.
In this work, we address these challenges by introducing a novel pipeline for multi-object sketch animation. We focus on vector sketches  as they do not require additional training data and offer better generalizability. 
Our analysis of existing works reveals that relying solely on the priors of a pre-trained T2V model is insufficient for achieving complex motions, mainly due to the model's limited ability to understand abstract sketches and associate implicit motion priors with specific target objects. 
Therefore, the essential idea of this work is to inject grouping and motion trajectory priors, beyond T2V priors, into the optimization process.

% More details of the novel components
Specifically, we propose a simple yet effective method called \textit{GroupSketch}, which consists of two main stages: \textbf{Motion Initialization} and \textbf{Motion Refinement}. 
In the \textit{first} stage, we design an interactive process that enables users to explicitly divide a sketch into groups and define key frames. A coarse dynamic sketch is then produced by automatically interpolating between these key frames, serving as initialization for the second stage.
In the \textit{second} stage, we introduce a novel Group-based Displacement Network (GDN) to refine the motion of the initial dynamic sketch. 
While this stage generally follows the LiveSketch pipeline, it incorporates several special design considerations for multi-object scenarios.
%by iteratively predicting the displacement field for each sketch group and optimizing with a pre-trained T2V model. 
Since GDN processes individual groups of a sketch and may overlook contextual information, we introduce a new component, Context-conditioned Feature Enhancement (CCFE), to inject context and further improve temporal consistency and overall animation quality. 
An SDS loss is then employed to distill the knowledge from a pre-trained T2V model to refine the motion of each group.

Our model can handle multi-object cases and realize complex motions through three critical designs: (1) Leveraging grouping priors. By dividing a sketch into stroke groups corresponding to semantic objects, GDN predicts displacement field for each group/object individually, thereby avoiding motion entanglement between different objects; (2) Two-stage pipeline. By incorporating motion trajectory priors, the model first generates a coarse animation, and initializing from this  animation effectively reduces the difficulty for GDN; (3) Special architectural designs in GDN to handle stroke groups of a sketch.
Extensive experiments have been conducted to validate the effectiveness of the proposed pipeline and individual components.
Our main contributions are summarized as follows:
\begin{itemize}[left=0pt, topsep=6pt, itemsep=4pt]
\item We propose a novel two-stage pipeline for vector sketch animation that effectively handles multi-object cases and complex motions, overcoming key limitations of existing approaches by integrating user-defined grouping and motion trajectory priors with T2V model capabilities.
\item We introduce a novel Group-based Displacement Network, where the Context-conditioned Feature Enhancement (CCFE) ensures temporal consistency and improves  animation quality by incorporating global context information.
\item Through extensive experimentation, we demonstrate that our method significantly outperforms existing sketch animation approaches in both visual quality and temporal consistency, particularly for complex multi-object scenarios.
%while maintaining better generalizability by operating on vector sketches without requiring additional training data.
% \item We design an interactive workflow that enables a more precise response to user inputs and adherence to textual prompts, ensuring generated animations meet user expectations.
\end{itemize}
%In summary, our approach supports heterogeneous transformation patterns, enabling more effective simulation of complex physical interactions in real-world scenarios and providing new solutions for the field of vector graphic animation generation.

\section{Related Work}
\subsection{Vector Sketch Generation}
Sketching is a fundamental tool for humans to express ideas~\cite{fan2023drawing, fan2018common, hertzmann2020line}. In recent years, significant progress has been made in the field of automatic sketch generation~\cite{xu2022deep}. Based on the representation methods, existing research can be categorized into pixel-based approaches and vector-based approaches. Pixel-based methods directly generate sketches in the image space~\cite{kampelmuhler2020synthesizing, li2019photo, song2018learning, xie2015holistically}, while vector-based methods produce structured representations composed of points, lines, and curves, which align more closely with the natural way humans draw sketches~\cite{bhunia2020pixelor, chen2017sketch, bhunia2022doodleformer,  ha2017neural, li2017free, lin2020sketch, mihai2021learning, mo2021general, ribeiro2020sketchformer}. A substantial body of sketch-based research has focused on vector graphics or SVG (Scalable Vector Graphics) synthesis, owing to their superior compatibility with visual design applications when compared to raster images~\cite{xing2024empowering, xing2024svgfusion, hu2024vectorpainter}. Traditional sketch generation methods typically rely on manually drawn sketch datasets for training. 
With the advancement of large-scale pre-trained language-vision models, recent studies~\cite{frans2022clipdraw, vinker2023clipascene, vinker2022clipasso, xing2024diffsketcher, xing2024svgdreamer, xing2025svgdreamer++} have begun leveraging the prior knowledge embedded in these models to reduce dependence on specialized sketch datasets. 
%Such approaches are capable of generating sketches in various styles or with different levels of abstraction.

\subsection{Sketch Animation}
Computer graphics has historically focused on creating user-friendly tools to transform static drawings into dynamic animations. In the realm of character animation, motion is commonly modeled as a sequence of time-varying poses, which are frequently described using user-defined inputs like stick figures~\cite{davis2006sketching}, skeletal structures~\cite{levi2013artisketch, pan2011sketch}, or segments of bone lines~\cite{oztireli2013differential}. And generate corresponding sketch animations from the given video~\cite{zheng2024sketch}.

Current text-driven sketch animation methods are primarily based on transfer learning from pre-trained diffusion models. Gal et al.~\cite{gal2024breathing} pioneered the LiveSketch framework, which achieved sketch animation generation without requiring additional training data. The core of this method lies in utilizing SDS~\cite{poole2023dreamfusion} to iteratively optimize vector animation sequences guided by the gradients of a pre-trained text-to-video (T2V) diffusion model. Subsequent research has advanced along two directions: on one hand, certain studies~\cite{rai2024enhancing, yang2024sketchanimator} have enhanced animation quality by advancing LiveSketch framework; on the other hand, Bandyopadhyay et al.~\cite{bandyopadhyay2024flipsketch} introduced DDIM inversion to process input sketches and adopted a fine-tuning strategy to optimize the parameters of the T2V model. However, these methods exhibit significant limitations: their motion modeling units are designed only for single objects, lacking a mechanism to model physical constraints for multi-object interactions, thereby restricting the generation performance in complex scenes.
Adopting the zero-shot learning paradigm of Live-Sketch, we employ SDS to harness the generative priors of a pre-trained text-to-video diffusion model for animation guidance, eliminating the need for task-specific training data.

\subsection{Image-to-Video Generation}
In recent years, text-guided image-to-video (I2V) generation~\cite{hu2024animate, tang2023any, wang2023videocomposer} technology has become a research hotspot in the field of computer vision. This technology generates video sequences that align with semantic content and exhibit specific motion patterns by combining static images with textual descriptions. Several innovative works have emerged in this research direction: Videocrafter~\cite{chen2023videocrafter1}, as the first open-source foundational model, achieves high-quality I2V generation while ensuring content consistency; I2VGenXL~\cite{zhang2023i2vgen} adopts a cascaded generation strategy, constructing low-resolution video frames first and then refining them into high-resolution outputs; DynamiCrafter~\cite{xing2024dynamicrafter} creatively leverages T2V prior knowledge and integrates a dual-stream image injection mechanism to generate dynamic content.
However, existing methods are primarily optimized for natural images. When applied to inputs like sketches, they still face severe challenges in maintaining structural consistency and motion rationality.

\begin{figure*}[t]
  \centering
  \includegraphics[width=\linewidth]{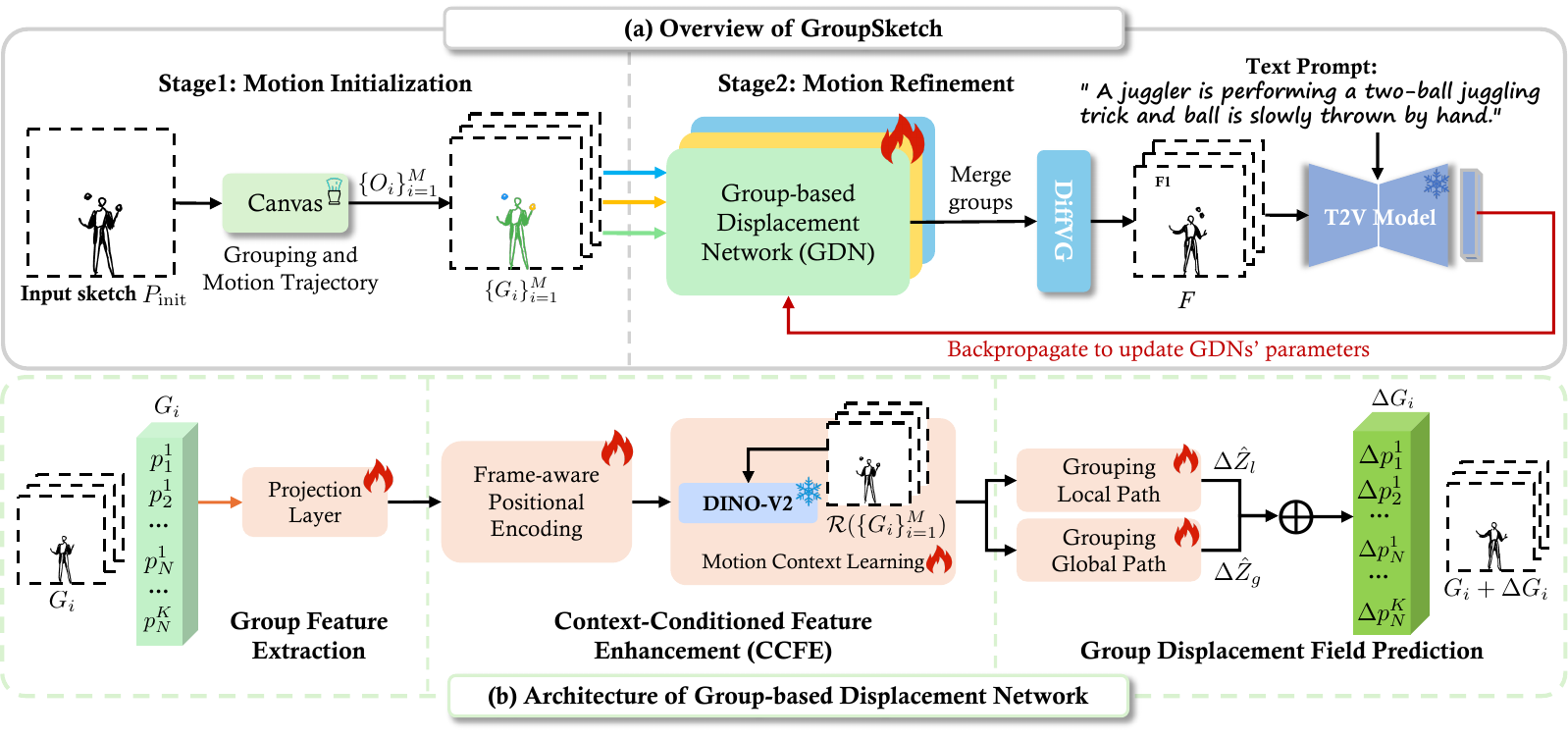}
  \vspace{-2.5em}
  \caption{Overview of the proposed method \textit{GroupSketch} and the architecture of the Group-based Displacement Network (GDN).
  \small (a) In the Motion Initialization Stage, the model takes an input sketch and obtains semantic groups and motion trajectories through a Canvas-based interactive process. This stage outputs a coarse-level sketch animation. In the Motion Refinement Stage, these groups are fed into GDN, which computes displacement fields to refine their motion. The updated motion is merged and then rendered by a differentiable rasterizer. The calculated loss is backpropagated to update GDN's parameters. (b) The GDN architecture consists of two components: (1) Context-conditioned Feature Enhancement (CCFE), which includes Frame-aware Positional Encoding (FPE) for encoding temporal positions of input point sequences and Motion Context Learning (MCL) for enhancing features with context information from all frames; (2) Group Displacement Field Prediction, which combines local and global grouping paths to produce final displacements for each group.}
  \Description{}
  % \vspace{-1em}
  \label{img:pipeline}
\end{figure*}

\begin{figure}[t]
\centering
\includegraphics[width=\linewidth]{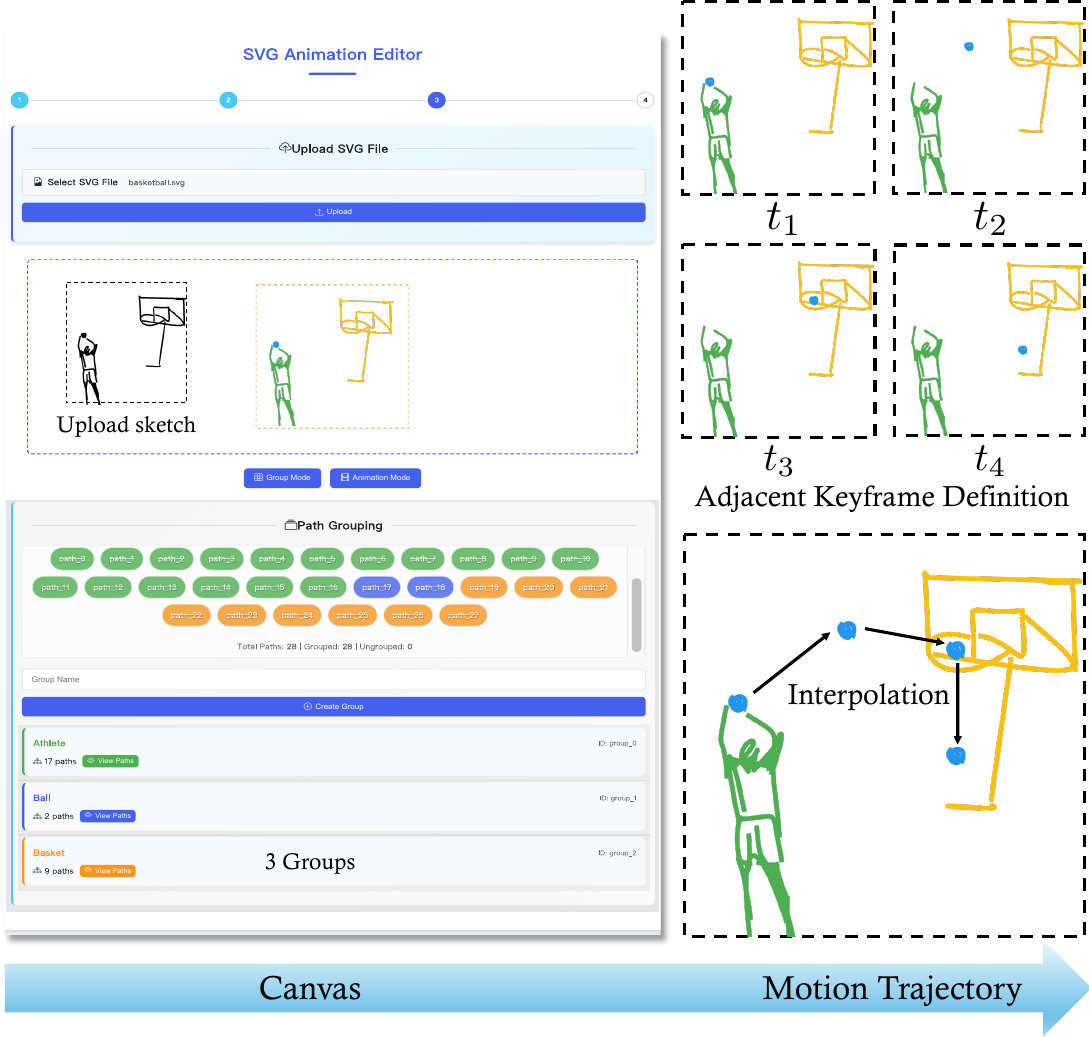}
\vspace{-2em}
\caption{Workflow of grouping and motion trajectory extraction, 
\small which users segment into semantic groups using an interactive Canvas. Users then customize keyframes and position each group in the Canvas. Finally, smooth motion trajectories are generated through interpolation between keyframes, enabling continuous and realistic motion of individual components in the scene.
}
\vspace{-1em}
\label{img:canvas}
\end{figure}

\section{Preliminaries}
\noindent \textbf{Sketch Representation with B\'ezier curves}.\quad 
The input vector sketch is represented as a collection of strokes rendered on a white canvas, where each stroke is modeled as a sequence of cubic Bézier curve segments.
%Keep in line with LiveSketch~\cite{gal2024breathing}, in the process of generating vector animations, we represent the input vector graphics as a set of two-dimensional Bézier curves, each curve defined by four control points. 
%Each control point is represented by its coordinates as $p = (x, y) \in \mathbb{R}^2$. We represent all the control points in a single frame as $P = \{p_1, p_2, \dots, p_N \}\in \mathbb{R}^{N\times 2}$, where $N$ represents the total number of control points in the input sketch, which remains constant in all generated frames. 
Each control point is defined by its coordinates $p = (x, y) \in \mathbb{R}^2$.
We denote the set of control points in a single frame as $P = \{p_1, p_2, \dots, p_N \}\in \mathbb{R}^{N\times 2}$, where $N$ represents the total number of control points in the input sketch $P_{init}$. 
To represent a video with $K$ frames, we define it as a sequence of $K$ sets of control points, denoted by $Z = \{P^k\}_{k=1}^K \in \mathbb{R}^{N \times K \times 2}$.
%We represent the set of control points in the initial sketch as $P_{init}$, and copy it $K$ times to create the initial frame sequence $Z_{init}=\left\{P_{init}^{j}\right\}_{j=1}^{K} \in \mathbb{R}^{K \times N \times 2}$.
Our goal is to transform this static sequence of frames into an animated sequence that reflects the motion described in a text prompt.
We formulate this task as learning a set of 2D displacements, $\Delta Z=\{ \Delta p_n^{k} \}_{n \in N}^{k \in K}$, where $\Delta p_n^{k}$ indicates the displacement of each control point $p_n$ in  frame $k$.
%We formulate this task as learning a set of 2D displacements $\Delta Z_{init}=\left\{\Delta P^{j}\right\}_{j=1}^{K} \in \mathbb{R}^{K \times N \times 2}$, indicating the displacement of each point $p_i^j$ in each frame $j$.

\noindent \textbf{Score Distillation Sampling}.
% DreamFusion~\cite{poole2023dreamfusion} proposed Score Distillation Sampling (SDS), which is a powerful optimization technique that distills the knowledge from pretrained diffusion models. This method allows a parametric model to refine its outputs using guidance from a text-conditioned diffusion prior.
DreamFusion~\cite{poole2023dreamfusion} introduced \emph{Score Distillation Sampling} (SDS), an iterative optimization technique that leverages pretrained diffusion models to train a differentiable generator $g$, which transforms learnable parameters $\theta$ into an image $\bm{x} = g(\theta)$.
The image is perturbed with noise based on a predefined diffusion schedule, resulting in a noised version at timestep $t$: $\bm{x}_t = \alpha_t \bm{x} + \sigma_t \epsilon$, where $\alpha_t$ and $\sigma_t$ are time-dependent scaling factors, and $\epsilon \sim \mathcal{N}(\bm{0}, \bm{I})$ denotes Gaussian noise.
Subsequently, the noised image $\bm{x}_t$ is processed by a pretrained diffusion model conditioned on a text prompt $y$ describing the target scene. 
The model predicts the noise component $\epsilon_\theta(\bm{x}_t; y, t)$, and the discrepancy between this prediction and the actual noise $\epsilon$ provides a gradient signal.
The gradient used to update the generator parameters is computed as:
\begin{equation}
\nabla_{\theta} \mathcal{L}_{\text{SDS}}(\phi, \bm{x}=g(\theta)) = \mathbb{E}_{t,\epsilon} \left[ 
w(t) (\epsilon_\theta(\bm{x}_t;y, t) - \epsilon ) \frac{\partial \bm{x}}{\partial \theta} 
\right]
\label{eq:sds_loss}
\end{equation}
where $w(t)$ is a weighting function determined by the diffusion schedule. Through iterative refinement, the generator converges to produce samples that are semantically aligned with the given textual description.

\noindent \textbf{Neural Displacement Field.}
The \textit{Neural Displacement Field} (NDF) introduced in LiveSketch~\cite{gal2024breathing} serves as the backbone for animating vector sketches. Inspired by Neural Radiance Fields (NeRF)~\cite{mildenhall2021nerf}, NDF learns per-point displacements by combining local and global transformations. It comprises three components: a shared backbone, a local pathway, and a global pathway.
% The neural displacement field in LiveSketch~\cite{gal2024breathing} serves as the backbone network, designed to learn per-point displacement vectors. 
% It consists of a shared backbone, a local pathway, and a global pathway. The shared backbone processes input control points to extract fundamental features.
%The shared backbone processes the input control points to extract fundamental features that are used across all frames. 
The shared backbone processes the input control points $P = \{p_1, \dots, p_N\} \in \mathbb{R}^{N \times 2}$ to extract fundamental features $\bm{F} = \text{MLP}(P)$, where $\text{MLP}(\cdot)$ is a feature extraction function shared across all frames.
%The local pathway employs a multilayer perceptron (MLP) to predict point-wise offsets, capturing fine-grained deformations for individual control points. In contrast, the global pathway predicts an affine transformation matrix that computes global displacements for all control points across frames. By combining local and global displacements, the NDF generates temporally coherent animations that align with the intended motion.
% Furthermore, the NDF leverages Score Distillation Sampling (SDS) loss~\cite{poole2023dreamfusion} to distill motion priors from pretrained text-to-video diffusion models. This ensures that the generated animations adhere closely to textual descriptions provided as prompts. The integration of local and global pathways with SDS-guided optimization allows LiveSketch to produce dynamic and visually consistent results.
The local pathway uses an MLP to predict point-wise displacements
$\Delta Z_l = \text{MLP}_{\text{local}}(\bm{F})$, where $\Delta Z_l \in \mathbb{R}^{N \times K \times 2}.$
The global pathway predicts an affine matrix $\bm{T}_g = \text{MLP}_{\text{global}}(\bm{F})$ via another $\text{MLP}_{\text{global}}(\cdot)$, and applies it to compute global motion $\Delta Z_g = \bm{T}_g \cdot P.$
The final displacement is the sum of both $\Delta Z = \Delta Z_l + \Delta Z_g.$
To align animations with textual prompts, NDF incorporates Score Distillation Sampling (SDS)~\cite{poole2023dreamfusion}, which distills motion priors from pretrained text-to-video diffusion models. The SDS loss jointly optimizes both pathways to ensure semantic alignment, sketch fidelity, and smooth motion.

\section{Methodology}
We propose \textit{GroupSketch}, a novel framework for multi-object vector sketch animations guided by natural language prompts. 
Our approach allows users to provide a static sketch together with a textual description of the intended motion, and interactively synthesizes temporally coherent animations that accurately capture both the semantics and dynamics specified by the input prompt.

\subsection{Overview of GroupSketch Pipeline}
As illustrated in Fig.~\ref{img:pipeline}, our method consists of two main stages: \textit{Motion Initialization} and \textit{Motion Refinement}. 
In the \textit{first stage}, we design an interactive process that allows users to explicitly divide a sketch into groups and define key frames. A coarse dynamic sketch is then generated by automatically interpolating between these key frames.
In the \textit{second} stage, we introduce a Group-based Displacement Network (GDN) to refine the motion of the coarse dynamic sketch, ultimately producing a final animated sketch that reflects the motion specified in the text prompt. 
%organizes the input sketch into motion-aware semantic groups, while the \textit{second stage} refines the motion trajectories through a dedicated displacement network and aligns the animation with the textual intent using differentiable rendering and language supervision.

\textit{Stage 1: Motion Initialization.} Given an input vector sketch $P_{\text{init}}$, we first perform grouping and motion trajectory initialization via an \textit{Interactive Canvas}. 
Specifically, we utilize user-provided motion trajectory information to compute per-frame displacement offsets for each semantic group. 
These offsets are integrated with the original sketch data $Z$, resulting in a set of motion-aware group representations $\{G_i\}_{i=1}^M$, where $M$ denotes the number of semantic groups and $G_i$ represents the sketch and trajectory data associated with the $i$-th group. This process establishes the foundation for disentangled motion modeling in the subsequent stage.

\textit{Stage 2: Motion Refinement.} The group representations $\{G_i\}_{i=1}^M$ are then fed into the \textit{Group-based Displacement Network} (GDN), which consists of $M$ parallel subnetworks.
%$\{\text{GDN}_i\}_{i=1}^M$. 
Each subnetwork independently models the motion dynamics of a specific group, thereby enabling structured decomposition of complex multi-object motion. 
The outputs of all subnetworks are merged and rendered as a sequence of $K$ frames using the differentiable renderer DiffVG~\cite{li2020differentiable}. 
To ensure semantic consistency with the user's textual description, we leverage a frozen T2V model~\cite{wang2023modelscope}, which evaluates the generated animation against the provided prompt and injects motion prior knowledge based on SDS loss~\cite{poole2023dreamfusion} to regulate intra-group and inter-group dynamics.

\subsection{Stage 1: Motion Initialization}
\label{subsec:motion_trajectory}
%Gradients from the T2V supervision are backpropagated through DiffVG, enabling fine-grained adjustment of SVG parameters and achieving alignment between visual motion and textual semantics.
%To enable interactive control over sketch animation, 
We introduce an \textit{Interactive Canvas} that allows users to specify object groupings, \textit{i.e.}, the group label for each stroke, and their associated motion trajectories, \textit{i.e.}, the trajectory for each group. 
The motion trajectory of each point is encoded as a basic 2D offset $\{ \Delta x, \Delta y \} \in \mathbb{R}^2$, and the set of offsets for a group across $K$ frames is denoted as $O = \{ o_k \}_{k=1}^{K}$. 
The complete trajectory representation across $M$ groups is then defined as $L = \left\{ O_i \right\}_{i=1}^{M}$.
%The basic offset in the motion trajectory data is represented as $ d = \left\{ \Delta x,\Delta y \right\}\in \mathbb{R}^2$, while $D = \left\{d_i \right\}_{i=1}^K$ denotes the set of offsets within the same group across $k$ frames. Finally, $ L = \left\{D_i \right\}_{i=1}^M$ encapsulates all the motion trajectory data.

%To empower users with interactive control over sketch grouping and motion trajectory design, we have developed an intuitive and user-friendly canvas-based method as shown in Figure~\ref{img:canvas}. 
As shown in Fig.~\ref{img:canvas}, using the Interactive Canvas, users can manually group strokes and define keyframes to provide the semantic and trajectory priors.  
Specifically, users select the strokes corresponding to an  object or a part and assign them a semantic label. Users then specify the spatial positions of a  semantic group through \textit{drag-and-drop} operations.
In Fig.~\ref{img:canvas}, four positions are indicated for the semantic group \textit{basketball}, corresponding to four keyframes. 
Finally, intermediate frames are interpolated between each pair of adjacent keyframes, resulting in a sketch sequence with coarse animation.
For example, given two adjacent keyframes \( t_1 \) and \( t_2 \), the intermediate frame \( t_i \) is linearly  interpolated following Eq.~\ref{eq:keyframe}.
\begin{equation}
\label{eq:keyframe}
% p(f)=(x_f,y_f)=p_{k_1} + t\cdot(p_{k_n} - p_{k_1})
e_{t_i} = (1-t) \cdot e_{t_1} + t \cdot e_{t_2}
\end{equation}
where $e_{t_i} = (x_{t_i}, y_{t_i})$ refer to the positions of control points in a keyframe $t_i$, $t=({t_i-t_1})/(t_2-t_1) \in [0,1]$ denotes the relative position of frame $t_i$ between keyframes $t_1$ and $t_2$. 
%This formula leverages linear interpolation to produce smooth transitional location data between two keyframes, offering a highly intuitive experience for users.
Note that we use the input sketch as the first keyframe while using the last keyframe as the final frame of the sketch sequence. This interpolation strategy ensures smooth transitions between spatial positions over time. 
%The resultant sketch sequence serves as the initialization for the second stage.
%This method enables users to tailor the motion patterns and trajectories of various objects in the animation based on their creative ideas and preferences, providing precise motion trajectory priors for the algorithm while transforming users into true sketch animation artists. 

\subsection{Stage 2: Motion Refinement}
\label{sec:group_based_displacement_network}
After the first stage, we obtain a  sketch sequence with coarse animation. 
To make the sketch animation more realistic, we propose a displacement prediction network  to refine the motion trajectory of each semantic group. As shown in Fig.~\ref{img:pipeline}(b), the \textbf{G}roup-aware \textbf{D}isplacement \textbf{N}etwork (GDN) is proposed to process \textit{individual} semantic group. Next, we will explain the architecture of GDN.

\subsubsection{Group Feature Extraction}
\label{sec:group_feature_extraction}

The Group Feature Extraction module functions as the initial embedding layer for each GDN component. %extracting the raw geometric information within a specific group. 
This module transforms the low-dimensional coordinate representation of each point into a higher-dimensional feature vector. 
Specifically, we use an MLP to transform \( G_i \in \mathbb{R}^{N \times K \times 2} \) into \( F^i_{pos} \in \mathbb{R}^{N \times K \times d} \), where $d$ is the hidden dimension.
% This transformation is achieved using a dedicated feed-forward network, referred to as the ``Projection Layer'' in our pipeline diagram (Fig.~\ref{img:pipeline}(b)). This network takes the 2D coordinates $(x, y)$ of each point as input and applies a series of non-linear transformations to map these coordinates into a richer, higher-dimensional feature space. The resulting feature vectors serve as the initial representation within the group, capturing essential geometric properties suitable for subsequent processing by the GDN.

\subsubsection{Context-conditioned Feature Enhancement (CCFE)}
\label{sec:group_based_context_condi_enhancement}
%解释encoding包含两部分：一部分是position encoding（intra-frame),还有temporal encoding (inter-frame).然后解释每一种encoding的运算过程。
This module consists of two components, Frame-aware Positional Encoding which incorporates both inter- and intra-frame positional information, and Motion Context Learning which injects visual context information into the vector embedding. 
%In this module, we add positional information to the animation vector embedding and fuse the visual supervision.

\noindent\textbf{Frame-aware Positional Encoding (FPE)}. \quad 
Each point in a sketch sequence has two positions, point-level and frame-level, corresponding to its \textit{spatial} position and \textit{temporal} position respectively. Therefore, we introduce FPE to perform point-level spatial encoding and frame-level temporal encoding. 

For spatial encoding, we employ standard trigonometric positional encoding~\cite{vaswani2017attention, gal2024breathing}, assigning different spatial position embeddings to different points within a frame.
For temporal encoding cross  frames, we assign the same temporal position embeddings to all points within a frame. 
To generate unique  representations for each frame, we employ a multi-frequency signal combination of sine and cosine functions at various frequencies.
%This design borrows from the concept of
Unlike the standard positional encoding, it specifically applies to the temporal dimension. 
Specifically, different dimensions of the encoding vector capture periodic signals across a spectrum of frequencies, from high to low. High-frequency components enable the model to distinguish between adjacent or temporally close frames, thereby capturing short-term dependencies. In contrast, low-frequency components span longer time windows, allowing the model to recognize long-term dependencies and overall motion trends.
This dual encoding strategy encourages the model to learn physically plausible motion patterns in vector animations by providing it with both precise spatial context and rich temporal dynamics.

% Finally, we add the temporal encoding with positional encoding. 
% The multi-frequency composition simultaneously captures both short-term and long-term temporal dependencies. 
% The design promotes physically plausible motion generation in vector animations.
%Combined with positional encoding, this mechanism enables precise spatio-temporal relationship modeling. 
%Enriched frequency features and nonlinear transformations facilitate learning complex temporal patterns. 

\noindent\textbf{Motion Context Learning}.\quad 
%原vector feature是从一个group的不同帧中提取得到的。然后注入全局信息，连续帧中提取的位图信息。先是编码连续帧的时序信息，然后再和vector group sketch feature融合。
% To effectively integrate the dual-modal features of vector graphics (vector paths and rendered images), we propose a Vector-Image Spatio-Temporal Attention mechanism that enables deep interaction between vector and visual features. Our approach consists of two core components: a temporal attention module for frame sequence modeling and a spatial cross-attention module for feature fusion.
%The integration of visual supervision and vector embedding will help enhance vector features~\cite{xing2024svgfusion}.
%Then, we employ DINOv2~\cite{oquab2023dinov2} to extract pixel-level features from the rendered images of each frame of the sketch after the initialization of the motion trajectory, denoted as $\bm{X} \in \mathbb{R}^{F \times P \times C}$, where $F$ represents the number of frames, $P$ indicates the number of patches, and $C$ is the feature dimension. After reshaping the features $\bm{X}$, we apply temporal attention across frame dimensions. This module is specifically designed to capture long-range temporal dependencies within frame sequences. 
To enhance the quality of vector representations, we inject visual supervision into vector embedding, as demonstrated in prior work~\cite{xing2024svgfusion}. Following the initialization of the motion trajectory, we leverage DINOv2~\cite{oquab2023dinov2} to extract pixel-level features from the rendered images $\mathcal{R}(\{G_i\}_{i=1}^M)$, which are rendered using DiffVG for each frame of the sketch sequence.
These features are denoted as $\bm{F_{bg}}$.
The extracted features $\bm{F_{bg}}$ are reshaped and subsequently processed using a temporal attention module designed to model long-range dependencies across the temporal dimension. This facilitates the learning of coherent motion patterns and temporal dynamics across frames.
For cross-modal fusion, we adopt a spatial cross-attention mechanism that uses vector features as queries and temporally-aware visual features as keys and values.  To further enhance spatio-temporal interaction, we employ a multi-layer attention framework based on stacked transformer blocks.
%Additional ablation experiments on DINOv2 are provided in Supp~\ref{supp:dinov2}. Detailed information about the CCFE architecture is provided in Supp.~\ref{supp:CCFE}.
%For cross-modal fusion, our spatial cross-attention mechanism enables deep integration between vector features and temporally-aware visual features. The architecture employs vector features as queries while using visual features as both keys and values. Following the cross-attention operation, we incorporate residual connections for stable gradient flow. 
%To achieve more comprehensive feature interaction, we implement a multi-layer spatio-temporal attention structure with stacked transformer blocks.

\subsubsection{Group Displacement Field Prediction}
Inspired by Neural Displacement Fields~\cite{gal2024breathing}, we propose a dual-path architecture for group-wise motion modeling. % after dual-modal feature fusion.
This architecture consists of a \textit{local path} and a \textit{global path}, designed to address both fine-grained deformations and object-level transformations.

\noindent\textbf{Grouping Local Path.}\quad  
The local path, parameterized by $\hat{M}_{l}$, uses a lightweight MLP to map the fused features to control point offsets $\Delta \hat{Z}_l$ for each point in group ${G}_i$, where $i$ indexes the group. This path captures localized motion and enables precise control point adjustments aligned with user-provided prompts.

\noindent\textbf{Grouping Global Path.} \quad  
Real-world scenarios often necessitate applying differentiated transformations to distinct objects, as shown in the example in Fig.~\ref{img:pipeline}.
%This limitation becomes particularly evident in contexts such as basketball scenes. For instance, applying a single, unified affine transformation prevents the basketball from rotating naturally, as the human body does not undergo identical rotational motion. Furthermore, subjecting both the basketball and the athlete's hand to the same translational transformation inhibits the separation required for realistic dribbling actions. Additionally, relying solely on the fine-grained local motion component proves insufficient for achieving coordinated movement among multiple objects.
Our proposed Grouping Global Path  enables the assignment of distinct affine transformation motions to different groups of objects, which facilitates multi-object motion decoupling and allows objects possessing varying physical properties or interaction dynamics to exhibit movement that is simultaneously independent and coordinated. 
The global path, parameterized by $\hat{M}_{g}$, predicts an affine transformation matrix $\bm{T}^k$ for each group at frame $k$, using the same fused features. 
This module captures global object motions, such as translation and rotation, enabling realistic and physically plausible behaviors (\textit{e.g.,} allowing the cats and fish to move independently, as shown in Fig.~\ref{img:compare}). The global displacement at frame $k$ is computed by applying $\bm{T}^k$ to each point $G_i^k$ and subtracting the original position:
$\Delta \hat{Z}_{\text{g}}^k = \bm{T}^k \cdot G_i^k - G_i^k$, where $\bm{T}^k$ is composed of scaling, shearing, rotation, and translation transformations applied sequentially.
% If the affine matrix is denoted as \( \bm{T}^j \), the global branch displacement for each point in the same group at frame \( j \) can be expressed as:
% \begin{equation}
% \Delta Z_{\text{gg}}^j = \bm{T}^j \bm{p}^{\text{init}} - \bm{p}^{\text{init}}
% \end{equation}
%In frame $j$, the global branch displacement for each point in the same group can be computed by transforming the point $\bm{G}_i^{j}$ using the affine transformation matrix $\bm{T}^j$ and then subtracting the original point \( \bm{G}_i^{j} \), resulting in the displacement $\Delta \hat Z_{\text{g}}^j$.

To enhance user control, we introduce scaling factors for each transformation component: $\lambda_t$, $\lambda_r$, $\lambda_s$, and $\lambda_{sh}$ for translation, rotation, scaling, and shear, respectively. 
For example, a predicted translation $(t_{jx}, t_{jy})$ is modulated as $(\lambda_t t_{jx}, \lambda_t t_{jy})$. 
This design enables fine-tuning of motion characteristics. 
Given that motion trajectory priors already provide coarse video-level translations, $\lambda_t$ can be set to a small default value for subtle adjustments. Users may further customize all transformation weights to achieve desired effects.
%motion dynamics.

%To enhance user control over generated motion, we introduce scaling factors for each type of transformation. Specifically, we set parameters $\lambda_t$, $\lambda_r$, $\lambda_s$, and $\lambda_{sh}$ for translation, rotation, scaling, and shear, respectively. For example, if the network predicts translation parameters $(t_{jx}, t_{jy})$, we adjust their magnitude by multiplying them by $\lambda_t$, resulting in new translation parameters $(\lambda_t t_{jx}, \lambda_t t_{jy})$. This method allows users to finely tune various aspects of the motion. Besides, since we have already obtained coarse video data through motion trajectory priors, we possess prior knowledge of translations. Thus, users can set the translation weight \( \lambda_t \) to a relatively small default value, only used for subtle translations. Users can also manually adjust the weights of scaling, shearing, rotation, and translation for each mode to achieve their desired animation. 

The final predicted displacement for group $i$ is the sum of the local and global branches: $\Delta G_i = \Delta \hat{Z}_{\text{l}} + \Delta \hat{Z}_{\text{g}}$.
This equation enables group-wise displacement prediction that balances detailed deformation and coherent global motion.
%delta G    K_k
%The final predicted displacement is the sum of the two branches: $\Delta \hat Z_{\text{l}} + \Delta \hat Z_{\text{g}}$, represented as $\Delta p_i$ for the $i$-th group, and we achieve group-wise prediction.

%\noindent\textbf{Joint Optimization SDS}. We merge all the groups and render them using DiffVG~\cite{li2020differentiable} into a sequence of frames containing all the sketch objects. The prediction network of the grouping realizes the combination of different motion patterns through the rendered sequence of frames and the prompt, and the optimization method of merging all groups realizes the necessary physical constraints between different motion modes. Finally, we apply the SDS~\cite{poole2023dreamfusion} loss to optimize the trainable parameters of each group.

\noindent\textbf{Joint Optimization with SDS.}\quad   
To enforce inter-group consistency and physical plausibility, we merge all groups and render a sequence of composite frames using DiffVG~\cite{li2020differentiable}, where all sketch objects are present. 
The grouping prediction network captures diverse motion patterns, while the joint rendering introduces necessary constraints between different motion modes. 
Finally, we apply the Score Distillation Sampling (SDS) loss~\cite{poole2023dreamfusion} to jointly optimize the trainable parameters of each group, guided by the rendered sequence and user prompts.

It is worth noting the hierarchy of motion guidance within our framework. The two sources of guidance, \textit{i.e.}, user-input motion trajectory priors from Stage 1 and T2V motion priors from Stage 2, are applied sequentially, with the framework primarily following the user-specified motion trajectories. The coarse animation from the first stage serves as a strong foundation for the second stage to further refine. Consequently, if there is a significant conflict between the user's input trajectory and the text prompt, 
the final animation will largely adhere to the user-defined motion but may exhibit unnatural effects. 
%This might result in a less natural final animation if the two guidance sources are misaligned.}

\section{Experiments}
\label{lab:results}
\subsection{Implementation Details}
\noindent\textbf{Training Details.}\quad We alternate between optimizing the Grouping Local Path and the Grouping Global Path, with all other trainable modules being optimized in both cases. We utilize Adam~\cite{kingma2014adam} as the optimizer, setting the learning rate for the Grouping Global Path to $1 \times 10^{-4}$ and for the Grouping Local Path to $5 \times 10^{-3}$. We set $\lambda_{\text{t}} = 1e-2$, $\lambda_{\text{r}} = 1e-2$, $\lambda_{\text{s}} = 5e-2$, and $\lambda_{\text{sh}} = 1e-1$. Similar to Livesketch, we employ ModelScope~\cite{wang2023modelscope} text-to-video as our diffusion backbone. We optimize all trainable modules for $500$ steps, with each video requiring approximately $40$ minutes on a V100 GPU and occupying about $22$ GB of memory. The hidden dimension $d$ of the entire model is set to $128$. We employ a 6-layer Vector-Image Spatio-Temporal Attention module. Note that adding a new semantic group increases memory usage by only approximately 150MB, indicating that memory is not a bottleneck for most use cases.

\noindent\textbf{Test Data Collection.}\quad Our test data collection methodology mirrors that of LiveSketch~\cite{gal2024breathing}. We utilized CLIPasso~\cite{vinker2022clipasso} to compile a multi-object vector sketch dataset comprising humans, animals, and inanimate objects, and instructed ChatGPT~\cite{ChatGPT} to randomly select 30 of these sketches along with prompts describing their characteristic motions. Additionally, from the LiveSketch dataset, we asked ChatGPT to randomly choose 15 single-object vector sketches and generate corresponding motion prompts for comparison.

\noindent\textbf{Baselines.}\quad The state-of-the-art methods for sketch animation generation are LiveSketch~\cite{vinker2022clipasso} and Flipsketch~\cite{bandyopadhyay2024flipsketch}, so we compare primarily with these two methods. To ensure the comprehensiveness of our experiments, we also compare with two text-guided image-to-video generation methods: (1) The I2VGen-XL~\cite{zhang2023i2vgen} model is a leading AI capable of synthesizing high-quality videos from images. It leverages a cascaded diffusion model to generate realistic videos. (2) DynamiCrafter~\cite{xing2024dynamicrafter} is a framework that uses pre-trained video diffusion priors and a dual-stream image injection mechanism to transform static images into realistic dynamic videos.

% The results of animation generation on multi-object sketches using our method are illustrated in Fig.~\ref{fig:teaser} and the videos in supplementary materials. Our approach successfully captures the dynamic behavior of various scenes, including a basketball player extending their arms to throw the ball, which rotates twice in the hoop before falling out; a diver and a fish swimming in water with rhythmic leg movements; and a juggler adjusting their posture to throw and catch balls. 
% For single-object sketch animations, as shown in Fig.~\ref{img:single}, our approach successfully captures the unfolding of an eagle in flight and the variations in shape of a surfboard at different angles, along with the diverse postures of a surfer. 
% Our method demonstrates an exceptional ability to capture complex actions and subtle changes, producing animations that are not only naturally fluid and realistic but also significantly enhanced in their artistic and expressive qualities, whether for single or multiple objects.
% By leveraging diverse text inputs and user interactions, our method can generate multiple different animation effects from the same sketch (as shown in Fig.~\ref{img:prompt}), which grants text-to-video technology greater creative freedom and expressiveness when handling abstract sketches.
\begin{figure}[t]
  \centering
  \includegraphics[width=\linewidth]{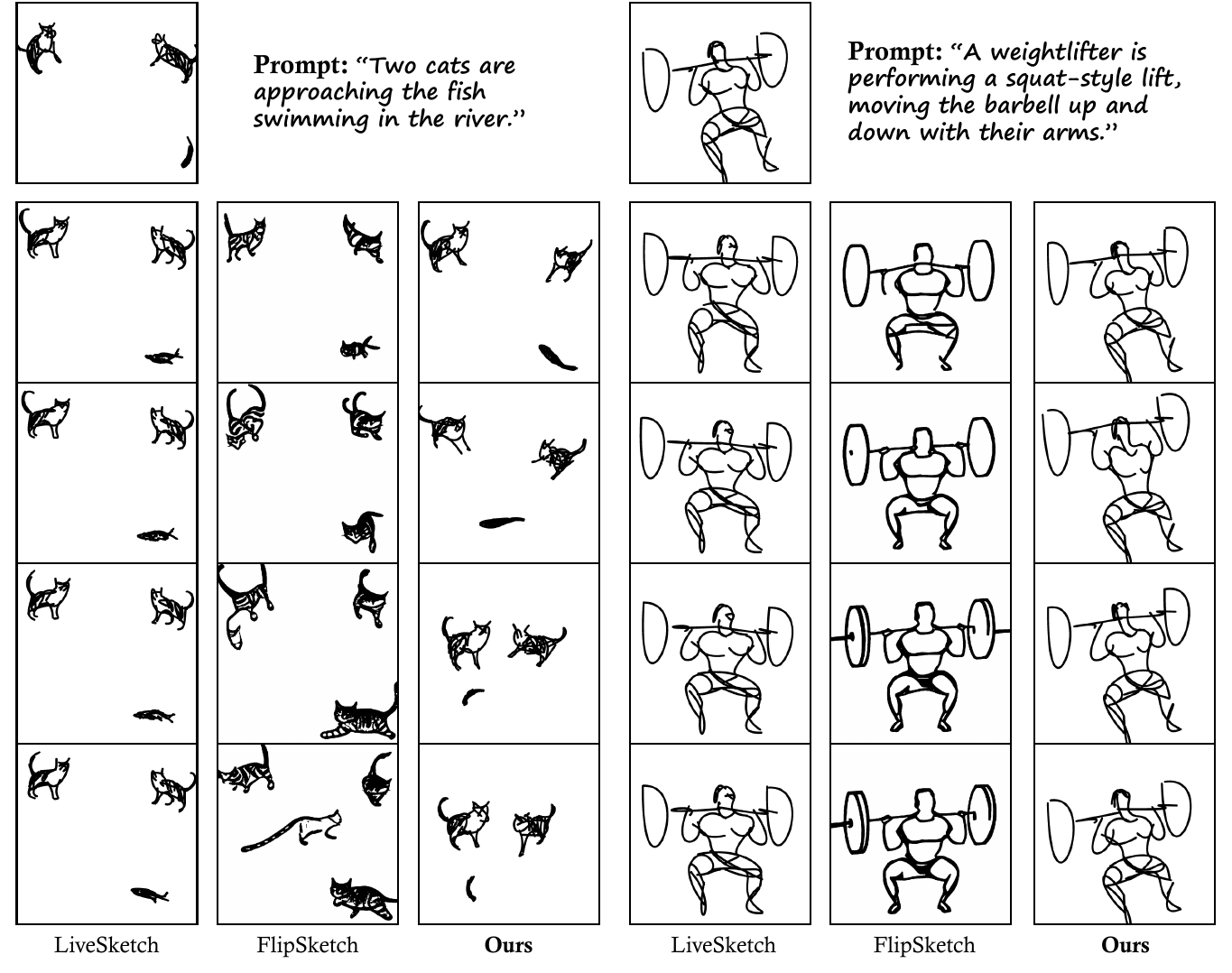}
  \vspace{-1em}
  \caption{Qualitative comparisons compared with FlipSketch and LiveSketch in the task of multi-object vector sketch animation synthesis. 
  \small FlipSketch struggles to maintain consistent visual features, while LiveSketch performs poorly in modeling complex motions, often resulting in animations with abnormal jittering or near-static behavior. In contrast, our proposed method effectively addresses these critical issues and demonstrates substantially superior animation performance.}
  \Description{}
  \vspace{-1em}
  \label{img:compare}
\end{figure}
\begin{figure}[t]
  \centering
  \includegraphics[width=\linewidth]{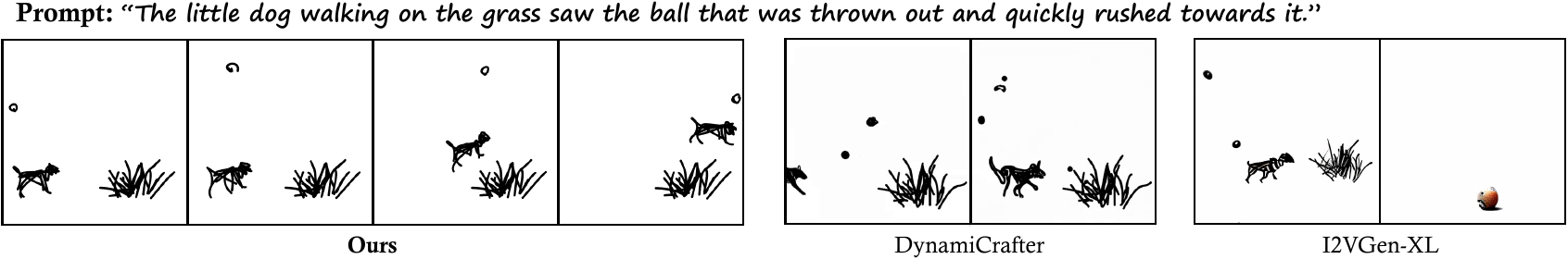}
  \vspace{-2em}
  \caption{Qualitative comparisons compared to Image-to-Video models. 
  \small Current I2V models exhibit significant artifacts and they struggle to maintain the original composition and fail to produce reasonable motion for sketch animation.}
  \Description{}
  \vspace{-1em}
  \label{img:i2v}
\end{figure}
\begin{figure}
  \centering
  \includegraphics[width=\linewidth]{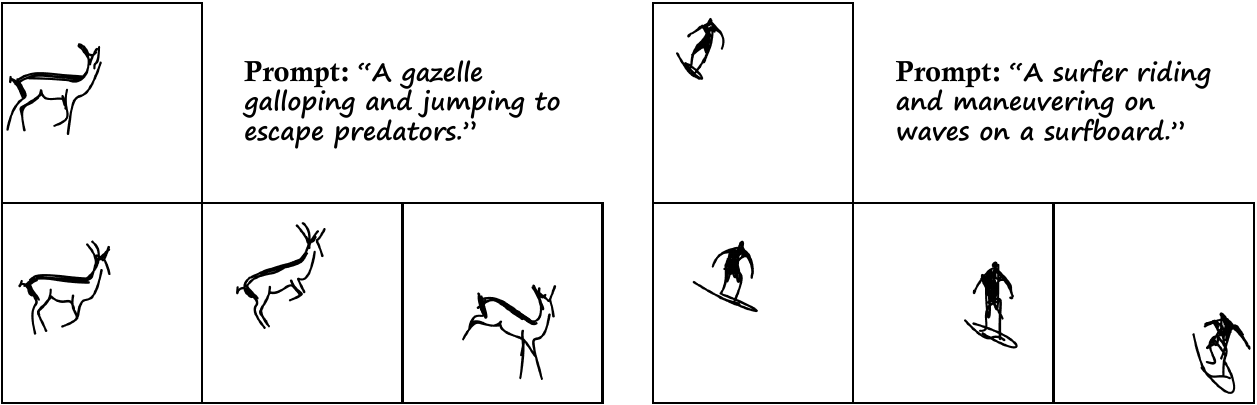}
  \vspace{-2em}
  \caption{Our method also excels in generating single-object sketch animations.}
  \Description{}
  \vspace{-1em}
  \label{img:single}
\end{figure}
\begin{figure}[t]
  \centering
  \includegraphics[width=\linewidth]{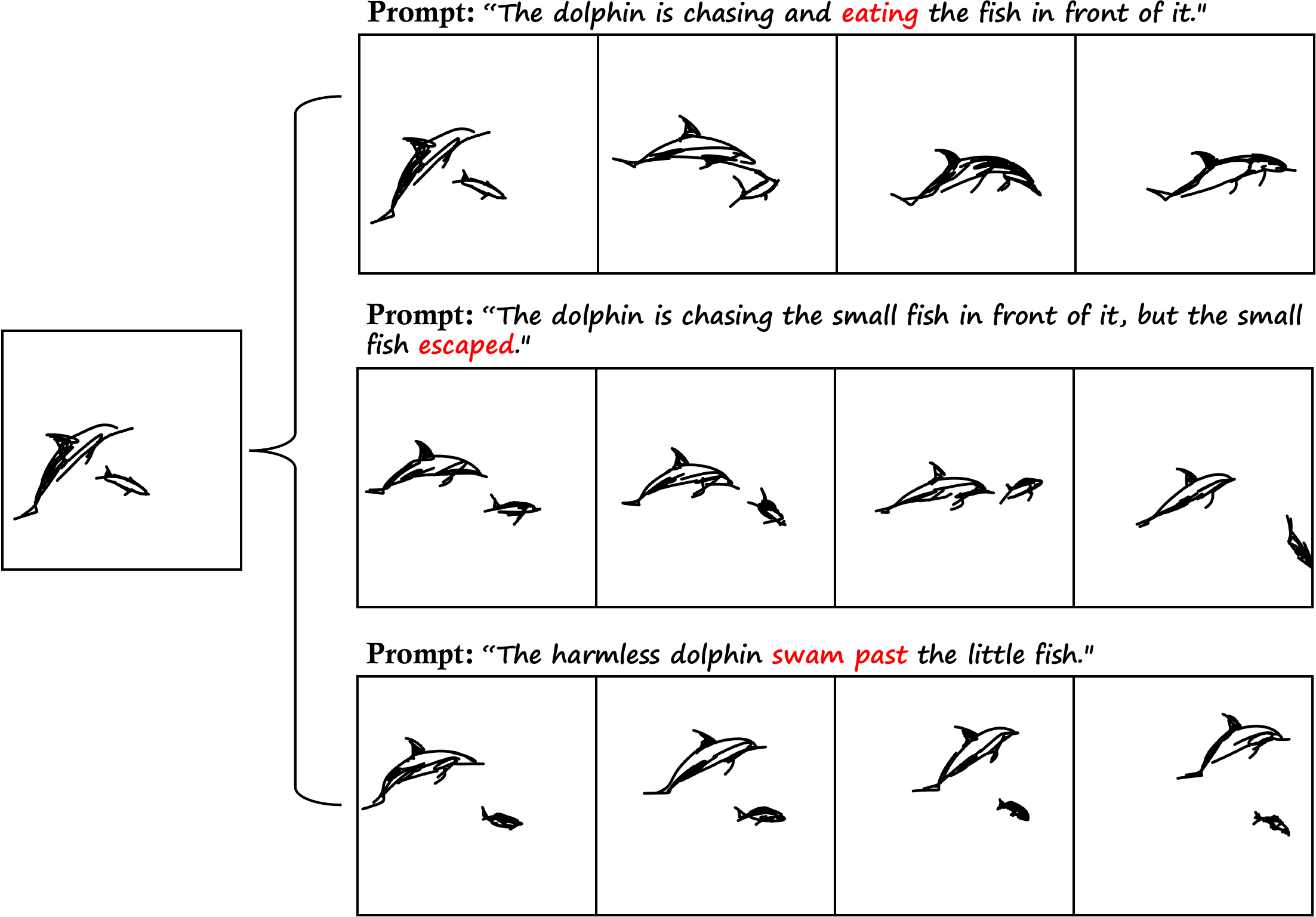}
  \vspace{-2em}
  \caption{Different prompts and user-defined motion trajectories can generate animations with different dynamic effects and semantics. 
  %The example at the bottom illustrates a significant conflict between user-defined trajectory (dolphin devours fish) and text prompt (fish escapes), resulting in an unnatural and disjointed animation where the fish abruptly breaks free in the final frames.
  % The safety of the little fish depends on you.
  }
  \Description{}
  % \vspace{-1em}
  \label{img:prompt}
\end{figure}

\noindent\textbf{Evaluation Metrics.}\quad
Consistent with prior work~\cite{vinker2022clipasso, bandyopadhyay2024flipsketch}, we evaluate performance using established metrics: (1) Sketch-to-Video Consistency: Measured using CLIP score~\cite{radford2021learning} between the input sketch and each generated video frame. (2) Text-to-Video Alignment: Assessed using X-CLIP~\cite{ni2022expanding}, which evaluates the semantic alignment between the generated video and the input text prompt.

\begin{figure*}[t]
  \centering
  \includegraphics[width=\linewidth]{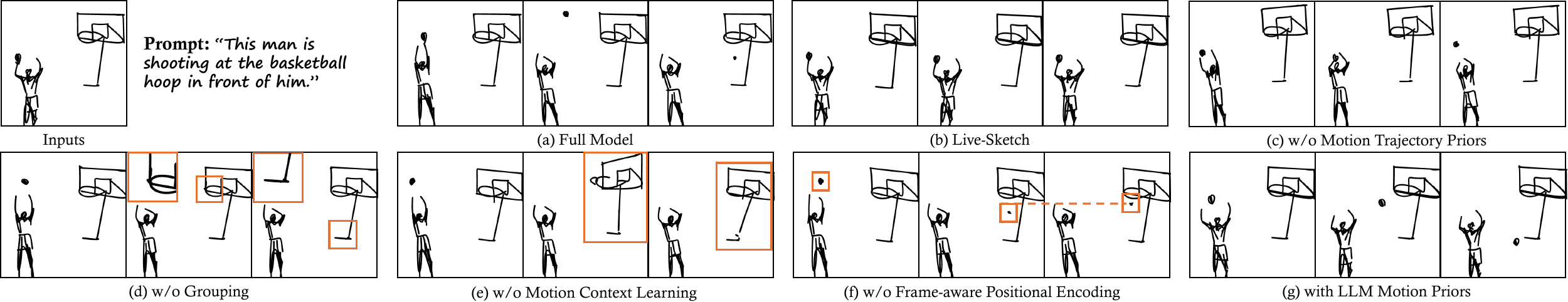}
  \vspace{-1em}
  \caption{Ablation studies of our proposed framework. 
  \small (a) Our full model, which generates harmonious motion of multi-object animation. (b) In the sketch animations generated by LiveSketch~\cite{gal2024breathing}, objects are largely stationary, with the basketball `stuck' to the player's hand, failing to achieve decoupling between objects. (c) Removing Motion Trajectory Priors results in inaccurate motion prediction, as the ball fails to reach the basket despite successful decoupling. (d) Without multi-object grouping, the strokes of the basketball merge with the basketball hoop, leading to a loss of object awareness and causing the basketball to disappear. (e) Removing Motion Context Learning leads to significant deformation of the basketball hoop, indicating impaired shape retention. (f) Without the FPE module, the basketball exhibits unnatural upward motion during descent, compromising temporal consistency. (g) The inability of current LLMs to accurately align the basketball's motion trajectory with the hoop, resulting in failed shots by the player in the animation, underscores the necessity of precise motion trajectories generated by the Canvas-based method.}
  \Description{}
  \label{img:ablation}
  % \vspace{-1em}
\end{figure*}

\subsection{Qualitative Results} 
Figure~\ref{img:compare} illustrates the qualitative comparison of our method with Livesketch and Flipsketch for multi-object sketch animation generation. Flipsketch's results lack temporal coherence, exhibiting abrupt transformations and artifacts, such as the disappearance of a fish followed by the appearance of two cats and a weightlifter failing to lift the barbell. Livesketch preserves the appearance of the sketches but often results in static animations when dealing with multiple objects because its global path applies affine transformation matrices to the entire sketch, causing objects to stick together. Our method generates a coarse video that is then refined through the GDN module by leveraging group decoupling and motion trajectory priors, effectively showcasing realistic interactions among different objects within the same animation (e.g., two cats in the river pouncing on a fish, while the fish quickly turns around and escapes).

To compare with Image-to-Video (I2V) models, we select DynamiCrafter and I2VGen-XL. However, as shown in Figure \ref{img:i2v}, these I2V models cannot maintain the shape of objects, leading to unreasonable outputs.

\begin{figure}[h]
  \centering
  \includegraphics[width=\linewidth]{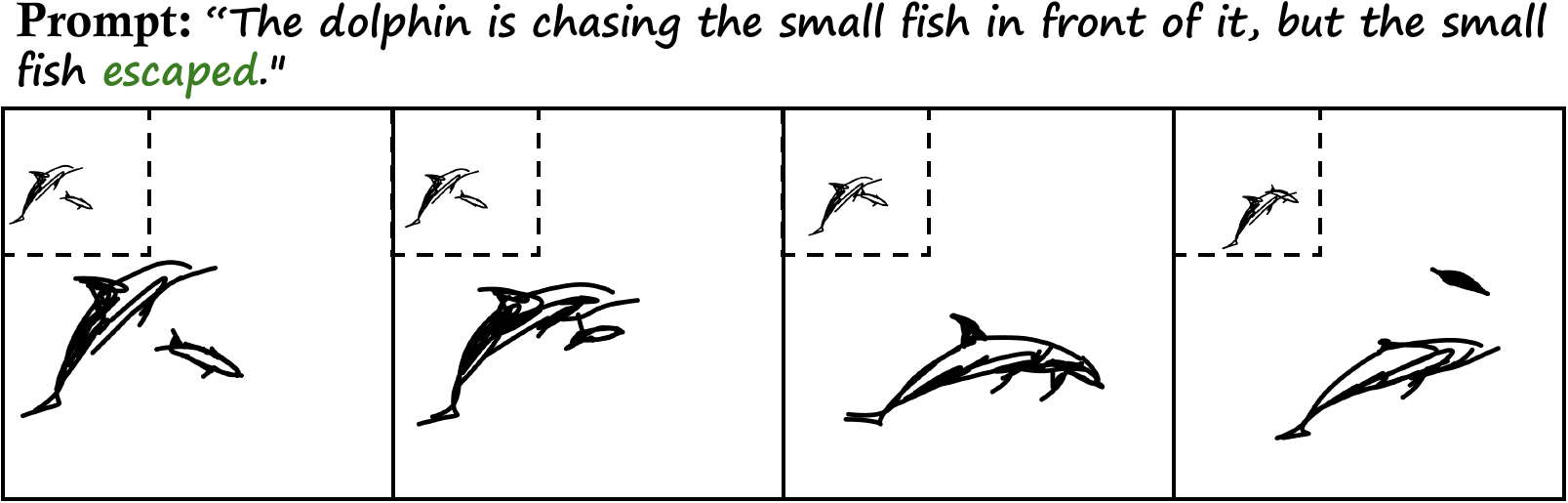}
  \vspace{-2em}
  \caption{An example illustrating a significant conflict between user-defined trajectory (dolphin devours fish) and text prompt (fish escaped), resulting in an unnatural and disjointed animation where the fish abruptly breaks free in the final frame.
  }
  \Description{}
  \vspace{-1em}
  \label{img:conflict}
\end{figure}

To further demonstrate the capabilities of our approach, we show more results of single- and multi-object sketch animation in Fig.~\ref{img:single} and Fig.~\ref{fig:teaser}. 
%the results of animation generation on multi-object sketches using our method are illustrated in Fig.~\ref{fig:teaser} and the project page. Our approach successfully captures the dynamic behavior of various scenes, including a basketball player extending their arms to throw the ball, which rotates twice in the hoop before falling out; a diver and a fish swimming in water with rhythmic leg movements; and a juggler adjusting their posture to throw and catch balls. 
%Similarly, for single-object sketch animations, as shown in Fig.~\ref{img:single}, our approach successfully captures the unfolding of an eagle in flight and the variations in shape of a surfboard at different angles, along with the diverse postures of a surfer. 
Overall, our method demonstrates an exceptional ability to capture complex actions and subtle changes, producing animations that are not only naturally fluid and realistic but also significantly enhanced in their artistic and expressive qualities, whether for single or multiple objects.
By leveraging diverse text inputs and user interactions, our method can generate multiple different animation effects from the same sketch, as shown in Fig.~\ref{img:prompt}. 
However, when there is a significant conflict between motion trajectory prior and text prompt, the resultant animation is likely to adhere to the user-defined motion trajectory, but with unnatural effects (Fig.~\ref{img:conflict}).
% which grants text-to-video technology greater creative freedom and expressiveness when handling abstract sketches.

%To evaluate the effectiveness of our method, we conducted comprehensive user studies covering qualitative animation results, grouping accuracy, and motion prediction. The detailed experimental setup, data, and results are presented in Supp.~\ref{supp:user_study}.
%----------------------------Table----------------------------
\begin{table}[t]
\centering 
\resizebox{\linewidth}{!}{
\begin{tabular}{cc}
    \begin{tabular}{lcc} 
        \toprule
        \multirow{2}{*}{Method} & Sketch-to-video & Text-to-Video  \\
         & consistency $\left(\uparrow\right)$ & alignment$\left(\uparrow\right)$ \\
        \midrule 
        I2VGen-XL & 0.764  & 0.113 \\
        Dynamicrafter & 0.797  & 0.122  \\
        Flipsketch & 0.864  & 0.154  \\
        Livesketch & 0.923  & 0.178  \\
        Ours & $\mathbf{0.945}$ & $\mathbf{0.226} $ \\
        \bottomrule \\
        \multicolumn{3}{c}{(a) Comparisons to baselines}
    \end{tabular}  &
    \begin{tabular}{lcc} 
        \toprule
        \multirow{2}{*}{Setup} & Sketch-to-video & Text-to-Video  \\
         & consistency $\left(\uparrow\right)$ & alignment$\left(\uparrow\right)$ \\
        \midrule 
        Full & 0.945 & 0.226 \\
        w/o Trajectory &  0.949 & 0.219  \\
        w/o Grouping&  0.941 &  0.221 \\
        w/o MCL &  0.938 &  0.222 \\
        w/o FPE&  0.944 &  0.224 \\
        \bottomrule \\
        \multicolumn{3}{c}{(b) Ablation results}
    \end{tabular}
\end{tabular}
}
\caption{Quantitative comparison of our method against baselines and ablation setups.}
\vspace{-2em}
\label{tab:all_quant}
\end{table}
\subsection{Quantitative Results}
Quantitative comparisons are reported in Table~\ref{tab:all_quant}(a).  Our approach performs better in both sketch-to-video consistency and text-to-video alignment. I2VGen-XL, DynamiCrafter, and FlipSketch achieve lower scores on both metrics due to their inability to follow complex instructions and maintain visual appearance. LiveSketch, represented in vector form, preserves visual appearance but struggles with generating complex motions, often resulting in static states. This limitation leads to lower scores in text-to-video alignment.

\subsection{Ablation Study}
\subsubsection{Component-Wise Ablation Analysis}
Qualitative and quantitative ablation studies are presented in Fig.~\ref{img:ablation} and Table~\ref{tab:all_quant}(b), respectively. We conducted multiple ablation experiments to demonstrate the effectiveness of our method. When the Motion Trajectory Priors were removed (Fig.~\ref{img:ablation}(c)), although the ball held by the athlete was successfully thrown, achieving decoupling compared to LiveSketch (Fig.~\ref{img:ablation}(b)), it failed to reach the basket. This indicates that the absence of trajectory priors compromises the precision of motion prediction. When the multi-object grouping operation was removed (Fig.~\ref{img:ablation}(d)), a noticeable phenomenon occurred: the strokes corresponding to the ball merged into the basketball hoop after entering the basket, resulting in the ``disappearance'' of the ball. This suggests that, without the grouping operation, our method loses its ability to perceive and distinguish multiple objects. When the Motion Context Learning was removed (Fig.~\ref{img:ablation}(e)), significant deformation of the basketball hoop was observed. This demonstrates that the network's visual perception capability for retaining shape information deteriorates in the absence of Motion Context Learning (MCL). Finally, when the Frame-aware Positional Encoding (FPE) module was removed (Fig.~\ref{img:ablation}(f)), the basketball exhibited an unnatural upward movement during its descent. This highlights that the network's ability to capture temporal sequences is weakened, leading to reduced temporal consistency. The results collectively underscore the importance of each component in our proposed framework and their contributions to the overall performance. %Furthermore, we conducted experiments to validate the effectiveness of the DINOv2~\cite{oquab2023dinov2} feature extractor employed in our approach. Detailed results and analyses of these experiments are presented in Supplementary Section \ref{supp:dinov2}.
\subsubsection{Ablation Study of Canvas-Based Method vs. LLM}
To further justify the inclusion of the interactive Canvas, we compared its effectiveness against using LLMs for automated prior generation. In our experiments, we utilized two of the most powerful multimodal large language models currently available, ChatGPT-4o and Claude 3.7 Sonnet, to automatically analyze SVG files and user-provided prompts. By highlighting individual strokes and grouping them based on their physical attributes and prompts, each group is assigned a unique color. Subsequently, LLMs generated motion trajectory data by reasoning about the expected physical relationships among grouped objects. 
This approach enabled the LLMs to provide motion predictions consistent with semantic context, serving as an alternative to the Canvas for generating animation prior knowledge.

Despite the promising capabilities of LLMs, our experiments revealed significant limitations when compared to Canvas-based interactive methods. In motion trajectory generation, LLMs exhibited imprecision in predicting movement patterns. In Figure~\ref{img:ablation}(g), the basketball initialized by LLMs failed to accurately pass through the hoop based on trajectory information.
In contrast, our Canvas-based method excelled in these tasks, providing users with precise control over grouping and trajectory design. 
This distinction establishes our Canvas-based method as a more reliable solution for generating high-quality vector animations.

%Through intuitive drag-and-drop operations on the Canvas interface, users could manually define keyframes for each object group, ensuring spatial accuracy across different time frames. Interpolation algorithms subsequently used linear interpolation to calculate intermediate positions, generating smooth and realistic motion trajectories. This interactive approach significantly reduced errors caused by ambiguous object boundaries and ensured temporal consistency.

%This comparative study highlights the advantages of our Canvas-based method in handling sketch grouping and motion trajectory generation tasks. While LLMs offer a degree of automation suitable for simple scenarios, their performance significantly declines in complex or abstract situations requiring fine-grained control. By integrating an interactive Canvas interface into our framework, users gain enhanced control over animation design, resulting in more accurate and visually coherent outcomes that align with creative intentions. This distinction establishes our Canvas-based method as a more reliable solution for generating high-quality vector animations.

\section{Conclusion}
%We propose a method with user-interaction capabilities that animates vector sketches according to textual prompts and applies them to the generation of multi-object sketch animations. Experimental results demonstrate that our approach effectively extracts motion patterns and physical constraints among multiple objects from the diffusion model, enabling object grouping and object-aware motion modeling. Through this innovation, we have successfully advanced the field of interactive multi-object sketch animation, opening up new avenues for future applications in sketch animation creation and interactive media.
We have proposed a novel approach for multi-object vector sketch animation. The essential idea of our approach is to leverage semantic groups and motion trajectories obtained through human interaction. Our two-stage pipeline first generates a coarse animation using semantic grouping and motion trajectories priors, and then refines it using motion priors from a pretrained text-to-video model. Extensive experiments have demonstrated the effectiveness of our proposed method.

\noindent\textbf{Limitation and Future Work.}\quad
Our approach has two main limitations. (1) The quality of the generated motion is constrained by the capabilities of the pretrained text-to-video (T2V) model. For motions unfamiliar to the T2V model, animation quality may degrade, or the generated motion may fail to adhere to the prompt. As the capabilities of T2V models improve, the quality of our animations will correspondingly enhance.
(2) Our method maintains a fixed number of strokes throughout the animation, which may lead to visual artifacts during complex interactions, such as occlusion. A potential solution is to introduce learnable opacity parameters for each stroke, allowing dynamic adjustment of visible strokes to create more realistic occlusion effects.

For future work, the manual Canvas-based interaction process could be accelerated by incorporating LLMs. Specifically, LLMs could be used to propose initial keyframes based on the text prompt, which users could then verify or adjust.
This could streamline the workflow while still retaining user control over the final results.
\begin{acks}
This work was supported in part by the Young Elite Scientists Sponsorship Program, sponsored by \grantsponsor{cast-YES}{Chinese Association for Science and Technology (CAST)}{https://english.cast.org.cn},
the \grantsponsor{centralfunds}{Fundamental Research Funds for the Central Universities}{https://www.cutech.edu.cn/en},
the \grantsponsor{NSFC}{National Natural Science Foundation of China}{https://doi.org/10.13039/501100001809} (~\grantnum{NSFC}{62461160331}, ~\grantnum{NSFC}{62132001}),
and the \grantsponsor{huaweibuaa}{Huawei-BUAA Joint Lab}{https://www.buaa.edu.cn}.
\end{acks}

\bibliographystyle{ACM-Reference-Format}
\bibliography{sample-base}


\begin{thebibliography}{58}

%%% the Web address.

\ifx \showCODEN    \undefined \def \showCODEN     #1{\unskip}     \fi
\ifx \showISBNx    \undefined \def \showISBNx     #1{\unskip}     \fi
\ifx \showISBNxiii \undefined \def \showISBNxiii  #1{\unskip}     \fi
\ifx \showISSN     \undefined \def \showISSN      #1{\unskip}     \fi
\ifx \showLCCN     \undefined \def \showLCCN      #1{\unskip}     \fi
\ifx \shownote     \undefined \def \shownote      #1{#1}          \fi
\ifx \showarticletitle \undefined \def \showarticletitle #1{#1}   \fi
\ifx \showURL      \undefined \def \showURL       {\relax}        \fi
% invisible to TeX
\providecommand\bibfield[2]{#2}
\providecommand\bibinfo[2]{#2}
\providecommand\natexlab[1]{#1}
\providecommand\showeprint[2][]{arXiv:#2}

\bibitem[Bandyopadhyay and Song(2025)]%
        {bandyopadhyay2024flipsketch}
\bibfield{author}{\bibinfo{person}{Hmrishav Bandyopadhyay} {and} \bibinfo{person}{Yi-Zhe Song}.} \bibinfo{year}{2025}\natexlab{}.
\newblock \showarticletitle{FlipSketch: Flipping Static Drawings to Text-Guided Sketch Animations}.
\newblock \bibinfo{journal}{\emph{Proceedings of the IEEE/CVF Conference on Computer Vision and Pattern Recognition}} (\bibinfo{year}{2025}).
\newblock


\bibitem[Berger et~al\mbox{.}(2013)]%
        {berger2013style}
\bibfield{author}{\bibinfo{person}{Itamar Berger}, \bibinfo{person}{Ariel Shamir}, \bibinfo{person}{Moshe Mahler}, \bibinfo{person}{Elizabeth Carter}, {and} \bibinfo{person}{Jessica Hodgins}.} \bibinfo{year}{2013}\natexlab{}.
\newblock \showarticletitle{Style and abstraction in portrait sketching}.
\newblock \bibinfo{journal}{\emph{ACM Transactions on Graphics}} \bibinfo{volume}{32}, \bibinfo{number}{4} (\bibinfo{year}{2013}), \bibinfo{pages}{1--12}.
\newblock


\bibitem[Bhunia et~al\mbox{.}(2020)]%
        {bhunia2020pixelor}
\bibfield{author}{\bibinfo{person}{Ayan~Kumar Bhunia}, \bibinfo{person}{Ayan Das}, \bibinfo{person}{Umar~Riaz Muhammad}, \bibinfo{person}{Yongxin Yang}, \bibinfo{person}{Timothy~M Hospedales}, \bibinfo{person}{Tao Xiang}, \bibinfo{person}{Yulia Gryaditskaya}, {and} \bibinfo{person}{Yi-Zhe Song}.} \bibinfo{year}{2020}\natexlab{}.
\newblock \showarticletitle{Pixelor: A Competitive Sketching AI Agent. So you think you can sketch?}
\newblock \bibinfo{journal}{\emph{ACM Transactions on Graphics (TOG)}} \bibinfo{volume}{39}, \bibinfo{number}{6} (\bibinfo{year}{2020}), \bibinfo{pages}{1--15}.
\newblock


\bibitem[Bhunia et~al\mbox{.}(2022)]%
        {bhunia2022doodleformer}
\bibfield{author}{\bibinfo{person}{Ankan~Kumar Bhunia}, \bibinfo{person}{Salman Khan}, \bibinfo{person}{Hisham Cholakkal}, \bibinfo{person}{Rao~Muhammad Anwer}, \bibinfo{person}{Fahad~Shahbaz Khan}, \bibinfo{person}{Jorma Laaksonen}, {and} \bibinfo{person}{Michael Felsberg}.} \bibinfo{year}{2022}\natexlab{}.
\newblock \showarticletitle{Doodleformer: Creative sketch drawing with transformers}. In \bibinfo{booktitle}{\emph{European Conference on Computer Vision}}. Springer, \bibinfo{pages}{338--355}.
\newblock


\bibitem[Chen et~al\mbox{.}(2023)]%
        {chen2023videocrafter1}
\bibfield{author}{\bibinfo{person}{Haoxin Chen}, \bibinfo{person}{Menghan Xia}, \bibinfo{person}{Yingqing He}, \bibinfo{person}{Yong Zhang}, \bibinfo{person}{Xiaodong Cun}, \bibinfo{person}{Shaoshu Yang}, \bibinfo{person}{Jinbo Xing}, \bibinfo{person}{Yaofang Liu}, \bibinfo{person}{Qifeng Chen}, \bibinfo{person}{Xintao Wang}, {et~al\mbox{.}}} \bibinfo{year}{2023}\natexlab{}.
\newblock \showarticletitle{Videocrafter1: Open diffusion models for high-quality video generation}.
\newblock \bibinfo{journal}{\emph{arXiv preprint arXiv:2310.19512}} (\bibinfo{year}{2023}).
\newblock


\bibitem[Chen et~al\mbox{.}(2017)]%
        {chen2017sketch}
\bibfield{author}{\bibinfo{person}{Yajing Chen}, \bibinfo{person}{Shikui Tu}, \bibinfo{person}{Yuqi Yi}, {and} \bibinfo{person}{Lei Xu}.} \bibinfo{year}{2017}\natexlab{}.
\newblock \showarticletitle{Sketch-pix2seq: a model to generate sketches of multiple categories}.
\newblock \bibinfo{journal}{\emph{arXiv preprint arXiv:1709.04121}} (\bibinfo{year}{2017}).
\newblock


\bibitem[Davis et~al\mbox{.}(2006)]%
        {davis2006sketching}
\bibfield{author}{\bibinfo{person}{James Davis}, \bibinfo{person}{Maneesh Agrawala}, \bibinfo{person}{Erika Chuang}, \bibinfo{person}{Zoran Popovi{\'c}}, {and} \bibinfo{person}{David Salesin}.} \bibinfo{year}{2006}\natexlab{}.
\newblock \showarticletitle{A sketching interface for articulated figure animation}.
\newblock In \bibinfo{booktitle}{\emph{Acm siggraph 2006 courses}}. \bibinfo{pages}{15--es}.
\newblock


\bibitem[Fan et~al\mbox{.}(2023)]%
        {fan2023drawing}
\bibfield{author}{\bibinfo{person}{Judith~E Fan}, \bibinfo{person}{Wilma~A Bainbridge}, \bibinfo{person}{Rebecca Chamberlain}, {and} \bibinfo{person}{Jeffrey~D Wammes}.} \bibinfo{year}{2023}\natexlab{}.
\newblock \showarticletitle{Drawing as a versatile cognitive tool}.
\newblock \bibinfo{journal}{\emph{Nature Reviews Psychology}} \bibinfo{volume}{2}, \bibinfo{number}{9} (\bibinfo{year}{2023}), \bibinfo{pages}{556--568}.
\newblock


\bibitem[Fan et~al\mbox{.}(2018)]%
        {fan2018common}
\bibfield{author}{\bibinfo{person}{Judith~E Fan}, \bibinfo{person}{Daniel~LK Yamins}, {and} \bibinfo{person}{Nicholas~B Turk-Browne}.} \bibinfo{year}{2018}\natexlab{}.
\newblock \showarticletitle{Common object representations for visual production and recognition}.
\newblock \bibinfo{journal}{\emph{Cognitive science}} \bibinfo{volume}{42}, \bibinfo{number}{8} (\bibinfo{year}{2018}), \bibinfo{pages}{2670--2698}.
\newblock


\bibitem[Frans et~al\mbox{.}(2022)]%
        {frans2022clipdraw}
\bibfield{author}{\bibinfo{person}{Kevin Frans}, \bibinfo{person}{Lisa Soros}, {and} \bibinfo{person}{Olaf Witkowski}.} \bibinfo{year}{2022}\natexlab{}.
\newblock \showarticletitle{Clipdraw: Exploring text-to-drawing synthesis through language-image encoders}.
\newblock \bibinfo{journal}{\emph{Advances in Neural Information Processing Systems}}  \bibinfo{volume}{35} (\bibinfo{year}{2022}), \bibinfo{pages}{5207--5218}.
\newblock


\bibitem[Gal et~al\mbox{.}(2024)]%
        {gal2024breathing}
\bibfield{author}{\bibinfo{person}{Rinon Gal}, \bibinfo{person}{Yael Vinker}, \bibinfo{person}{Yuval Alaluf}, \bibinfo{person}{Amit Bermano}, \bibinfo{person}{Daniel Cohen-Or}, \bibinfo{person}{Ariel Shamir}, {and} \bibinfo{person}{Gal Chechik}.} \bibinfo{year}{2024}\natexlab{}.
\newblock \showarticletitle{Breathing life into sketches using text-to-video priors}. In \bibinfo{booktitle}{\emph{Proceedings of the IEEE/CVF Conference on Computer Vision and Pattern Recognition}}. \bibinfo{pages}{4325--4336}.
\newblock


\bibitem[Ha and Eck(2017)]%
        {ha2017neural}
\bibfield{author}{\bibinfo{person}{David Ha} {and} \bibinfo{person}{Douglas Eck}.} \bibinfo{year}{2017}\natexlab{}.
\newblock \showarticletitle{A neural representation of sketch drawings}.
\newblock \bibinfo{journal}{\emph{arXiv preprint arXiv:1704.03477}} (\bibinfo{year}{2017}).
\newblock


\bibitem[Ha and Eck(2018)]%
        {ha2018a}
\bibfield{author}{\bibinfo{person}{David Ha} {and} \bibinfo{person}{Douglas Eck}.} \bibinfo{year}{2018}\natexlab{}.
\newblock \showarticletitle{A Neural Representation of Sketch Drawings}. In \bibinfo{booktitle}{\emph{International Conference on Learning Representations}}.
\newblock
\urldef\tempurl%
\url{https://openreview.net/forum?id=Hy6GHpkCW}
\showURL{%
\tempurl}


\bibitem[Hertzmann(2020)]%
        {hertzmann2020line}
\bibfield{author}{\bibinfo{person}{Aaron Hertzmann}.} \bibinfo{year}{2020}\natexlab{}.
\newblock \showarticletitle{Why do line drawings work? a realism hypothesis}.
\newblock \bibinfo{journal}{\emph{Perception}} \bibinfo{volume}{49}, \bibinfo{number}{4} (\bibinfo{year}{2020}), \bibinfo{pages}{439--451}.
\newblock


\bibitem[Hu et~al\mbox{.}(2025)]%
        {hu2024vectorpainter}
\bibfield{author}{\bibinfo{person}{Juncheng Hu}, \bibinfo{person}{Ximing Xing}, \bibinfo{person}{Jing Zhang}, {and} \bibinfo{person}{Qian Yu}.} \bibinfo{year}{2025}\natexlab{}.
\newblock \showarticletitle{VectorPainter: Advanced Stylized Vector Graphics Synthesis Using Stroke-Style Priors}. In \bibinfo{booktitle}{\emph{2025 IEEE International Conference on Multimedia and Expo (ICME)}}. IEEE, \bibinfo{pages}{1--6}.
\newblock


\bibitem[Hu(2024)]%
        {hu2024animate}
\bibfield{author}{\bibinfo{person}{Li Hu}.} \bibinfo{year}{2024}\natexlab{}.
\newblock \showarticletitle{Animate anyone: Consistent and controllable image-to-video synthesis for character animation}. In \bibinfo{booktitle}{\emph{Proceedings of the IEEE/CVF Conference on Computer Vision and Pattern Recognition}}. \bibinfo{pages}{8153--8163}.
\newblock


\bibitem[Kampelmuhler and Pinz(2020)]%
        {kampelmuhler2020synthesizing}
\bibfield{author}{\bibinfo{person}{Moritz Kampelmuhler} {and} \bibinfo{person}{Axel Pinz}.} \bibinfo{year}{2020}\natexlab{}.
\newblock \showarticletitle{Synthesizing human-like sketches from natural images using a conditional convolutional decoder}. In \bibinfo{booktitle}{\emph{Proceedings of the IEEE/CVF winter conference on applications of computer vision}}. \bibinfo{pages}{3203--3211}.
\newblock


\bibitem[Kingma and Ba(2014)]%
        {kingma2014adam}
\bibfield{author}{\bibinfo{person}{Diederik~P Kingma} {and} \bibinfo{person}{Jimmy Ba}.} \bibinfo{year}{2014}\natexlab{}.
\newblock \showarticletitle{s c}.
\newblock \bibinfo{journal}{\emph{arXiv preprint arXiv:1412.6980}} (\bibinfo{year}{2014}).
\newblock


\bibitem[Levi and Gotsman(2013)]%
        {levi2013artisketch}
\bibfield{author}{\bibinfo{person}{Zohar Levi} {and} \bibinfo{person}{Craig Gotsman}.} \bibinfo{year}{2013}\natexlab{}.
\newblock \showarticletitle{Artisketch: A system for articulated sketch modeling}. In \bibinfo{booktitle}{\emph{Computer Graphics Forum}}, Vol.~\bibinfo{volume}{32}. Wiley Online Library, \bibinfo{pages}{235--244}.
\newblock


\bibitem[Li et~al\mbox{.}(2019)]%
        {li2019photo}
\bibfield{author}{\bibinfo{person}{Mengtian Li}, \bibinfo{person}{Zhe Lin}, \bibinfo{person}{Radomir Mech}, \bibinfo{person}{Ersin Yumer}, {and} \bibinfo{person}{Deva Ramanan}.} \bibinfo{year}{2019}\natexlab{}.
\newblock \showarticletitle{Photo-sketching: Inferring contour drawings from images}. In \bibinfo{booktitle}{\emph{2019 IEEE Winter Conference on Applications of Computer Vision (WACV)}}. IEEE, \bibinfo{pages}{1403--1412}.
\newblock


\bibitem[Li et~al\mbox{.}(2020)]%
        {li2020differentiable}
\bibfield{author}{\bibinfo{person}{Tzu-Mao Li}, \bibinfo{person}{Michal Luk{\'a}{\v{c}}}, \bibinfo{person}{Micha{\"e}l Gharbi}, {and} \bibinfo{person}{Jonathan Ragan-Kelley}.} \bibinfo{year}{2020}\natexlab{}.
\newblock \showarticletitle{Differentiable vector graphics rasterization for editing and learning}.
\newblock \bibinfo{journal}{\emph{ACM Transactions on Graphics (TOG)}} \bibinfo{volume}{39}, \bibinfo{number}{6} (\bibinfo{year}{2020}), \bibinfo{pages}{1--15}.
\newblock


\bibitem[Li et~al\mbox{.}(2017)]%
        {li2017free}
\bibfield{author}{\bibinfo{person}{Yi Li}, \bibinfo{person}{Yi-Zhe Song}, \bibinfo{person}{Timothy~M Hospedales}, {and} \bibinfo{person}{Shaogang Gong}.} \bibinfo{year}{2017}\natexlab{}.
\newblock \showarticletitle{Free-hand sketch synthesis with deformable stroke models}.
\newblock \bibinfo{journal}{\emph{International Journal of Computer Vision}}  \bibinfo{volume}{122} (\bibinfo{year}{2017}), \bibinfo{pages}{169--190}.
\newblock


\bibitem[Lin et~al\mbox{.}(2020)]%
        {lin2020sketch}
\bibfield{author}{\bibinfo{person}{Hangyu Lin}, \bibinfo{person}{Yanwei Fu}, \bibinfo{person}{Xiangyang Xue}, {and} \bibinfo{person}{Yu-Gang Jiang}.} \bibinfo{year}{2020}\natexlab{}.
\newblock \showarticletitle{Sketch-bert: Learning sketch bidirectional encoder representation from transformers by self-supervised learning of sketch gestalt}. In \bibinfo{booktitle}{\emph{Proceedings of the IEEE/CVF Conference on Computer Vision and Pattern Recognition}}. \bibinfo{pages}{6758--6767}.
\newblock


\bibitem[Mihai and Hare(2021)]%
        {mihai2021learning}
\bibfield{author}{\bibinfo{person}{Daniela Mihai} {and} \bibinfo{person}{Jonathon Hare}.} \bibinfo{year}{2021}\natexlab{}.
\newblock \showarticletitle{Learning to draw: Emergent communication through sketching}.
\newblock \bibinfo{journal}{\emph{Advances in neural information processing systems}}  \bibinfo{volume}{34} (\bibinfo{year}{2021}), \bibinfo{pages}{7153--7166}.
\newblock


\bibitem[Mildenhall et~al\mbox{.}(2021)]%
        {mildenhall2021nerf}
\bibfield{author}{\bibinfo{person}{Ben Mildenhall}, \bibinfo{person}{Pratul~P Srinivasan}, \bibinfo{person}{Matthew Tancik}, \bibinfo{person}{Jonathan~T Barron}, \bibinfo{person}{Ravi Ramamoorthi}, {and} \bibinfo{person}{Ren Ng}.} \bibinfo{year}{2021}\natexlab{}.
\newblock \showarticletitle{Nerf: Representing scenes as neural radiance fields for view synthesis}.
\newblock \bibinfo{journal}{\emph{Commun. ACM}} \bibinfo{volume}{65}, \bibinfo{number}{1} (\bibinfo{year}{2021}), \bibinfo{pages}{99--106}.
\newblock


\bibitem[Mo et~al\mbox{.}(2021)]%
        {mo2021general}
\bibfield{author}{\bibinfo{person}{Haoran Mo}, \bibinfo{person}{Edgar Simo-Serra}, \bibinfo{person}{Chengying Gao}, \bibinfo{person}{Changqing Zou}, {and} \bibinfo{person}{Ruomei Wang}.} \bibinfo{year}{2021}\natexlab{}.
\newblock \showarticletitle{General virtual sketching framework for vector line art}.
\newblock \bibinfo{journal}{\emph{ACM Transactions on Graphics (TOG)}} \bibinfo{volume}{40}, \bibinfo{number}{4} (\bibinfo{year}{2021}), \bibinfo{pages}{1--14}.
\newblock


\bibitem[Ni et~al\mbox{.}(2022)]%
        {ni2022expanding}
\bibfield{author}{\bibinfo{person}{Bolin Ni}, \bibinfo{person}{Houwen Peng}, \bibinfo{person}{Minghao Chen}, \bibinfo{person}{Songyang Zhang}, \bibinfo{person}{Gaofeng Meng}, \bibinfo{person}{Jianlong Fu}, \bibinfo{person}{Shiming Xiang}, {and} \bibinfo{person}{Haibin Ling}.} \bibinfo{year}{2022}\natexlab{}.
\newblock \showarticletitle{Expanding language-image pretrained models for general video recognition}. In \bibinfo{booktitle}{\emph{European conference on computer vision}}. Springer, \bibinfo{pages}{1--18}.
\newblock


\bibitem[OpenAI(2023)]%
        {ChatGPT}
\bibfield{author}{\bibinfo{person}{OpenAI}.} \bibinfo{year}{2023}\natexlab{}.
\newblock \bibinfo{title}{Introducing ChatGPT}.
\newblock \bibinfo{howpublished}{\url{https://openai.com/index/chatgpt/}}.
\newblock


\bibitem[Oquab et~al\mbox{.}(2023)]%
        {oquab2023dinov2}
\bibfield{author}{\bibinfo{person}{Maxime Oquab}, \bibinfo{person}{Timoth{\'e}e Darcet}, \bibinfo{person}{Th{\'e}o Moutakanni}, \bibinfo{person}{Huy Vo}, \bibinfo{person}{Marc Szafraniec}, \bibinfo{person}{Vasil Khalidov}, \bibinfo{person}{Pierre Fernandez}, \bibinfo{person}{Daniel Haziza}, \bibinfo{person}{Francisco Massa}, \bibinfo{person}{Alaaeldin El-Nouby}, {et~al\mbox{.}}} \bibinfo{year}{2023}\natexlab{}.
\newblock \showarticletitle{Dinov2: Learning robust visual features without supervision}.
\newblock \bibinfo{journal}{\emph{arXiv preprint arXiv:2304.07193}} (\bibinfo{year}{2023}).
\newblock


\bibitem[{\"O}ztireli et~al\mbox{.}(2013)]%
        {oztireli2013differential}
\bibfield{author}{\bibinfo{person}{A~Cengiz {\"O}ztireli}, \bibinfo{person}{Ilya Baran}, \bibinfo{person}{Tiberiu Popa}, \bibinfo{person}{Boris Dalstein}, \bibinfo{person}{Robert~W Sumner}, {and} \bibinfo{person}{Markus Gross}.} \bibinfo{year}{2013}\natexlab{}.
\newblock \showarticletitle{Differential blending for expressive sketch-based posing}. In \bibinfo{booktitle}{\emph{Proceedings of the 12th ACM SIGGRAPH/Eurographics Symposium on Computer Animation}}. \bibinfo{pages}{155--164}.
\newblock


\bibitem[Pan and Zhang(2011)]%
        {pan2011sketch}
\bibfield{author}{\bibinfo{person}{Junjun Pan} {and} \bibinfo{person}{Jian~J Zhang}.} \bibinfo{year}{2011}\natexlab{}.
\newblock \bibinfo{booktitle}{\emph{Sketch-based skeleton-driven 2D animation and motion capture}}.
\newblock \bibinfo{publisher}{Springer}.
\newblock


\bibitem[Poole et~al\mbox{.}(2023)]%
        {poole2023dreamfusion}
\bibfield{author}{\bibinfo{person}{Ben Poole}, \bibinfo{person}{Ajay Jain}, \bibinfo{person}{Jonathan~T. Barron}, {and} \bibinfo{person}{Ben Mildenhall}.} \bibinfo{year}{2023}\natexlab{}.
\newblock \showarticletitle{DreamFusion: Text-to-3D using 2D Diffusion}. In \bibinfo{booktitle}{\emph{The Eleventh International Conference on Learning Representations (ICLR)}}.
\newblock


\bibitem[Qu et~al\mbox{.}(2023)]%
        {qu2023sketchdreamer}
\bibfield{author}{\bibinfo{person}{Zhiyu Qu}, \bibinfo{person}{Tao Xiang}, {and} \bibinfo{person}{Yi-Zhe Song}.} \bibinfo{year}{2023}\natexlab{}.
\newblock \showarticletitle{SketchDreamer: Interactive Text-Augmented Creative Sketch Ideation}. In \bibinfo{booktitle}{\emph{BMVC}}.
\newblock


\bibitem[Radford et~al\mbox{.}(2021)]%
        {radford2021learning}
\bibfield{author}{\bibinfo{person}{Alec Radford}, \bibinfo{person}{Jong~Wook Kim}, \bibinfo{person}{Chris Hallacy}, \bibinfo{person}{Aditya Ramesh}, \bibinfo{person}{Gabriel Goh}, \bibinfo{person}{Sandhini Agarwal}, \bibinfo{person}{Girish Sastry}, \bibinfo{person}{Amanda Askell}, \bibinfo{person}{Pamela Mishkin}, \bibinfo{person}{Jack Clark}, {et~al\mbox{.}}} \bibinfo{year}{2021}\natexlab{}.
\newblock \showarticletitle{Learning transferable visual models from natural language supervision}. In \bibinfo{booktitle}{\emph{International conference on machine learning}}. PmLR, \bibinfo{pages}{8748--8763}.
\newblock


\bibitem[Rai and Sharma(2024)]%
        {rai2024enhancing}
\bibfield{author}{\bibinfo{person}{Gaurav Rai} {and} \bibinfo{person}{Ojaswa Sharma}.} \bibinfo{year}{2024}\natexlab{}.
\newblock \showarticletitle{Enhancing Sketch Animation: Text-to-Video Diffusion Models with Temporal Consistency and Rigidity Constraints}.
\newblock \bibinfo{journal}{\emph{arXiv preprint arXiv:2411.19381}} (\bibinfo{year}{2024}).
\newblock


\bibitem[Ribeiro et~al\mbox{.}(2020)]%
        {ribeiro2020sketchformer}
\bibfield{author}{\bibinfo{person}{Leo Sampaio~Ferraz Ribeiro}, \bibinfo{person}{Tu Bui}, \bibinfo{person}{John Collomosse}, {and} \bibinfo{person}{Moacir Ponti}.} \bibinfo{year}{2020}\natexlab{}.
\newblock \showarticletitle{Sketchformer: Transformer-based representation for sketched structure}. In \bibinfo{booktitle}{\emph{Proceedings of the IEEE/CVF conference on computer vision and pattern recognition}}. \bibinfo{pages}{14153--14162}.
\newblock


\bibitem[Song et~al\mbox{.}(2021)]%
        {song2021ddim}
\bibfield{author}{\bibinfo{person}{Jiaming Song}, \bibinfo{person}{Chenlin Meng}, {and} \bibinfo{person}{Stefano Ermon}.} \bibinfo{year}{2021}\natexlab{}.
\newblock \showarticletitle{Denoising Diffusion Implicit Models}. In \bibinfo{booktitle}{\emph{International Conference on Learning Representations (ICLR)}}.
\newblock


\bibitem[Song et~al\mbox{.}(2018)]%
        {song2018learning}
\bibfield{author}{\bibinfo{person}{Jifei Song}, \bibinfo{person}{Kaiyue Pang}, \bibinfo{person}{Yi-Zhe Song}, \bibinfo{person}{Tao Xiang}, {and} \bibinfo{person}{Timothy~M Hospedales}.} \bibinfo{year}{2018}\natexlab{}.
\newblock \showarticletitle{Learning to sketch with shortcut cycle consistency}. In \bibinfo{booktitle}{\emph{Proceedings of the IEEE conference on computer vision and pattern recognition}}. \bibinfo{pages}{801--810}.
\newblock


\bibitem[Tang et~al\mbox{.}(2023)]%
        {tang2023any}
\bibfield{author}{\bibinfo{person}{Zineng Tang}, \bibinfo{person}{Ziyi Yang}, \bibinfo{person}{Chenguang Zhu}, \bibinfo{person}{Michael Zeng}, {and} \bibinfo{person}{Mohit Bansal}.} \bibinfo{year}{2023}\natexlab{}.
\newblock \showarticletitle{Any-to-any generation via composable diffusion}.
\newblock \bibinfo{journal}{\emph{Advances in Neural Information Processing Systems}}  \bibinfo{volume}{36} (\bibinfo{year}{2023}), \bibinfo{pages}{16083--16099}.
\newblock


\bibitem[Vaswani et~al\mbox{.}(2017)]%
        {vaswani2017attention}
\bibfield{author}{\bibinfo{person}{Ashish Vaswani}, \bibinfo{person}{Noam Shazeer}, \bibinfo{person}{Niki Parmar}, \bibinfo{person}{Jakob Uszkoreit}, \bibinfo{person}{Llion Jones}, \bibinfo{person}{Aidan~N Gomez}, \bibinfo{person}{{\L}ukasz Kaiser}, {and} \bibinfo{person}{Illia Polosukhin}.} \bibinfo{year}{2017}\natexlab{}.
\newblock \showarticletitle{Attention is all you need}.
\newblock \bibinfo{journal}{\emph{Advances in neural information processing systems}}  \bibinfo{volume}{30} (\bibinfo{year}{2017}).
\newblock


\bibitem[Vinker et~al\mbox{.}(2023)]%
        {vinker2023clipascene}
\bibfield{author}{\bibinfo{person}{Yael Vinker}, \bibinfo{person}{Yuval Alaluf}, \bibinfo{person}{Daniel Cohen-Or}, {and} \bibinfo{person}{Ariel Shamir}.} \bibinfo{year}{2023}\natexlab{}.
\newblock \showarticletitle{Clipascene: Scene sketching with different types and levels of abstraction}. In \bibinfo{booktitle}{\emph{Proceedings of the IEEE/CVF International Conference on Computer Vision}}. \bibinfo{pages}{4146--4156}.
\newblock


\bibitem[Vinker et~al\mbox{.}(2022)]%
        {vinker2022clipasso}
\bibfield{author}{\bibinfo{person}{Yael Vinker}, \bibinfo{person}{Ehsan Pajouheshgar}, \bibinfo{person}{Jessica~Y Bo}, \bibinfo{person}{Roman~Christian Bachmann}, \bibinfo{person}{Amit~Haim Bermano}, \bibinfo{person}{Daniel Cohen-Or}, \bibinfo{person}{Amir Zamir}, {and} \bibinfo{person}{Ariel Shamir}.} \bibinfo{year}{2022}\natexlab{}.
\newblock \showarticletitle{Clipasso: Semantically-aware object sketching}.
\newblock \bibinfo{journal}{\emph{ACM Transactions on Graphics (TOG)}} \bibinfo{volume}{41}, \bibinfo{number}{4} (\bibinfo{year}{2022}), \bibinfo{pages}{1--11}.
\newblock


\bibitem[Vinker et~al\mbox{.}(2024)]%
        {vinker2024sketchagent}
\bibfield{author}{\bibinfo{person}{Yael Vinker}, \bibinfo{person}{Tamar~Rott Shaham}, \bibinfo{person}{Kristine Zheng}, \bibinfo{person}{Alex Zhao}, \bibinfo{person}{Judith~E Fan}, {and} \bibinfo{person}{Antonio Torralba}.} \bibinfo{year}{2024}\natexlab{}.
\newblock \showarticletitle{SketchAgent: Language-Driven Sequential Sketch Generation}.
\newblock \bibinfo{journal}{\emph{arXiv preprint arXiv:2411.17673}} (\bibinfo{year}{2024}).
\newblock


\bibitem[Wang et~al\mbox{.}(2023a)]%
        {wang2023modelscope}
\bibfield{author}{\bibinfo{person}{Jiuniu Wang}, \bibinfo{person}{Hangjie Yuan}, \bibinfo{person}{Dayou Chen}, \bibinfo{person}{Yingya Zhang}, \bibinfo{person}{Xiang Wang}, {and} \bibinfo{person}{Shiwei Zhang}.} \bibinfo{year}{2023}\natexlab{a}.
\newblock \showarticletitle{Modelscope text-to-video technical report}.
\newblock \bibinfo{journal}{\emph{arXiv preprint arXiv:2308.06571}} (\bibinfo{year}{2023}).
\newblock


\bibitem[Wang et~al\mbox{.}(2023b)]%
        {wang2023videocomposer}
\bibfield{author}{\bibinfo{person}{Xiang Wang}, \bibinfo{person}{Hangjie Yuan}, \bibinfo{person}{Shiwei Zhang}, \bibinfo{person}{Dayou Chen}, \bibinfo{person}{Jiuniu Wang}, \bibinfo{person}{Yingya Zhang}, \bibinfo{person}{Yujun Shen}, \bibinfo{person}{Deli Zhao}, {and} \bibinfo{person}{Jingren Zhou}.} \bibinfo{year}{2023}\natexlab{b}.
\newblock \showarticletitle{Videocomposer: Compositional video synthesis with motion controllability}.
\newblock \bibinfo{journal}{\emph{Advances in Neural Information Processing Systems}}  \bibinfo{volume}{36} (\bibinfo{year}{2023}), \bibinfo{pages}{7594--7611}.
\newblock


\bibitem[Xie and Tu(2015)]%
        {xie2015holistically}
\bibfield{author}{\bibinfo{person}{Saining Xie} {and} \bibinfo{person}{Zhuowen Tu}.} \bibinfo{year}{2015}\natexlab{}.
\newblock \showarticletitle{Holistically-nested edge detection}. In \bibinfo{booktitle}{\emph{Proceedings of the IEEE international conference on computer vision}}. \bibinfo{pages}{1395--1403}.
\newblock


\bibitem[Xing et~al\mbox{.}(2024c)]%
        {xing2024dynamicrafter}
\bibfield{author}{\bibinfo{person}{Jinbo Xing}, \bibinfo{person}{Menghan Xia}, \bibinfo{person}{Yong Zhang}, \bibinfo{person}{Haoxin Chen}, \bibinfo{person}{Wangbo Yu}, \bibinfo{person}{Hanyuan Liu}, \bibinfo{person}{Gongye Liu}, \bibinfo{person}{Xintao Wang}, \bibinfo{person}{Ying Shan}, {and} \bibinfo{person}{Tien-Tsin Wong}.} \bibinfo{year}{2024}\natexlab{c}.
\newblock \showarticletitle{Dynamicrafter: Animating open-domain images with video diffusion priors}. In \bibinfo{booktitle}{\emph{European Conference on Computer Vision}}. Springer, \bibinfo{pages}{399--417}.
\newblock


\bibitem[Xing et~al\mbox{.}(2025a)]%
        {xing2024empowering}
\bibfield{author}{\bibinfo{person}{Ximing Xing}, \bibinfo{person}{Juncheng Hu}, \bibinfo{person}{Guotao Liang}, \bibinfo{person}{Jing Zhang}, \bibinfo{person}{Dong Xu}, {and} \bibinfo{person}{Qian Yu}.} \bibinfo{year}{2025}\natexlab{a}.
\newblock \showarticletitle{Empowering LLMs to Understand and Generate Complex Vector Graphics}.
\newblock \bibinfo{journal}{\emph{Proceedings of the IEEE/CVF Conference on Computer Vision and Pattern Recognition}} (\bibinfo{year}{2025}).
\newblock


\bibitem[Xing et~al\mbox{.}(2024a)]%
        {xing2024svgfusion}
\bibfield{author}{\bibinfo{person}{Ximing Xing}, \bibinfo{person}{Juncheng Hu}, \bibinfo{person}{Jing Zhang}, \bibinfo{person}{Dong Xu}, {and} \bibinfo{person}{Qian Yu}.} \bibinfo{year}{2024}\natexlab{a}.
\newblock \showarticletitle{SVGFusion: Scalable Text-to-SVG Generation via Vector Space Diffusion}.
\newblock \bibinfo{journal}{\emph{arXiv preprint arXiv:2412.10437}} (\bibinfo{year}{2024}).
\newblock


\bibitem[Xing et~al\mbox{.}(2024b)]%
        {xing2024diffsketcher}
\bibfield{author}{\bibinfo{person}{Ximing Xing}, \bibinfo{person}{Chuang Wang}, \bibinfo{person}{Haitao Zhou}, \bibinfo{person}{Jing Zhang}, \bibinfo{person}{Qian Yu}, {and} \bibinfo{person}{Dong Xu}.} \bibinfo{year}{2024}\natexlab{b}.
\newblock \showarticletitle{Diffsketcher: Text guided vector sketch synthesis through latent diffusion models}.
\newblock \bibinfo{journal}{\emph{Advances in Neural Information Processing Systems}}  \bibinfo{volume}{36} (\bibinfo{year}{2024}).
\newblock


\bibitem[Xing et~al\mbox{.}(2025b)]%
        {xing2025svgdreamer++}
\bibfield{author}{\bibinfo{person}{Ximing Xing}, \bibinfo{person}{Qian Yu}, \bibinfo{person}{Chuang Wang}, \bibinfo{person}{Haitao Zhou}, \bibinfo{person}{Jing Zhang}, {and} \bibinfo{person}{Dong Xu}.} \bibinfo{year}{2025}\natexlab{b}.
\newblock \showarticletitle{SVGDreamer++: Advancing Editability and Diversity in Text-Guided SVG Generation}.
\newblock \bibinfo{journal}{\emph{IEEE Transactions on Pattern Analysis and Machine Intelligence}} (\bibinfo{year}{2025}).
\newblock


\bibitem[Xing et~al\mbox{.}(2024d)]%
        {xing2024svgdreamer}
\bibfield{author}{\bibinfo{person}{Ximing Xing}, \bibinfo{person}{Haitao Zhou}, \bibinfo{person}{Chuang Wang}, \bibinfo{person}{Jing Zhang}, \bibinfo{person}{Dong Xu}, {and} \bibinfo{person}{Qian Yu}.} \bibinfo{year}{2024}\natexlab{d}.
\newblock \showarticletitle{Svgdreamer: Text guided svg generation with diffusion model}. In \bibinfo{booktitle}{\emph{Proceedings of the IEEE/CVF Conference on Computer Vision and Pattern Recognition}}. \bibinfo{pages}{4546--4555}.
\newblock


\bibitem[Xu et~al\mbox{.}(2022)]%
        {xu2022deep}
\bibfield{author}{\bibinfo{person}{Peng Xu}, \bibinfo{person}{Timothy~M Hospedales}, \bibinfo{person}{Qiyue Yin}, \bibinfo{person}{Yi-Zhe Song}, \bibinfo{person}{Tao Xiang}, {and} \bibinfo{person}{Liang Wang}.} \bibinfo{year}{2022}\natexlab{}.
\newblock \showarticletitle{Deep learning for free-hand sketch: A survey}.
\newblock \bibinfo{journal}{\emph{IEEE transactions on pattern analysis and machine intelligence}} \bibinfo{volume}{45}, \bibinfo{number}{1} (\bibinfo{year}{2022}), \bibinfo{pages}{285--312}.
\newblock


\bibitem[Yang et~al\mbox{.}(2024)]%
        {yang2024sketchanimator}
\bibfield{author}{\bibinfo{person}{Ruolin Yang}, \bibinfo{person}{Da Li}, \bibinfo{person}{Honggang Zhang}, {and} \bibinfo{person}{Yi-Zhe Song}.} \bibinfo{year}{2024}\natexlab{}.
\newblock \showarticletitle{SketchAnimator: Animate Sketch via Motion Customization of Text-to-Video Diffusion Models}. In \bibinfo{booktitle}{\emph{2024 IEEE International Conference on Visual Communications and Image Processing (VCIP)}}. IEEE, \bibinfo{pages}{1--5}.
\newblock


\bibitem[Yu et~al\mbox{.}(2016)]%
        {yu2016sketch}
\bibfield{author}{\bibinfo{person}{Qian Yu}, \bibinfo{person}{Feng Liu}, \bibinfo{person}{Yi-Zhe Song}, \bibinfo{person}{Tao Xiang}, \bibinfo{person}{Timothy~M Hospedales}, {and} \bibinfo{person}{Chen-Change Loy}.} \bibinfo{year}{2016}\natexlab{}.
\newblock \showarticletitle{Sketch me that shoe}. In \bibinfo{booktitle}{\emph{Proceedings of the IEEE conference on computer vision and pattern recognition}}. \bibinfo{pages}{799--807}.
\newblock


\bibitem[Yu et~al\mbox{.}(2017)]%
        {yu2017sketch}
\bibfield{author}{\bibinfo{person}{Qian Yu}, \bibinfo{person}{Yongxin Yang}, \bibinfo{person}{Feng Liu}, \bibinfo{person}{Yi-Zhe Song}, \bibinfo{person}{Tao Xiang}, {and} \bibinfo{person}{Timothy~M Hospedales}.} \bibinfo{year}{2017}\natexlab{}.
\newblock \showarticletitle{Sketch-a-net: A deep neural network that beats humans}.
\newblock \bibinfo{journal}{\emph{International journal of computer vision}}  \bibinfo{volume}{122} (\bibinfo{year}{2017}), \bibinfo{pages}{411--425}.
\newblock


\bibitem[Zhang et~al\mbox{.}(2023)]%
        {zhang2023i2vgen}
\bibfield{author}{\bibinfo{person}{Shiwei Zhang}, \bibinfo{person}{Jiayu Wang}, \bibinfo{person}{Yingya Zhang}, \bibinfo{person}{Kang Zhao}, \bibinfo{person}{Hangjie Yuan}, \bibinfo{person}{Zhiwu Qin}, \bibinfo{person}{Xiang Wang}, \bibinfo{person}{Deli Zhao}, {and} \bibinfo{person}{Jingren Zhou}.} \bibinfo{year}{2023}\natexlab{}.
\newblock \showarticletitle{I2vgen-xl: High-quality image-to-video synthesis via cascaded diffusion models}.
\newblock \bibinfo{journal}{\emph{arXiv preprint arXiv:2311.04145}} (\bibinfo{year}{2023}).
\newblock


\bibitem[Zheng et~al\mbox{.}(2024)]%
        {zheng2024sketch}
\bibfield{author}{\bibinfo{person}{Yudian Zheng}, \bibinfo{person}{Xiaodong Cun}, \bibinfo{person}{Menghan Xia}, {and} \bibinfo{person}{Chi-Man Pun}.} \bibinfo{year}{2024}\natexlab{}.
\newblock \showarticletitle{Sketch Video Synthesis}. In \bibinfo{booktitle}{\emph{Computer Graphics Forum}}, Vol.~\bibinfo{volume}{43}. Wiley Online Library, \bibinfo{pages}{e15044}.
\newblock


\end{thebibliography}

\end{document}